\documentclass{article} 
\usepackage{iclr2025_conference,times}
\usepackage{xspace}
\usepackage{mathpazo}
\usepackage{float}

\usepackage{amsmath,amsfonts,bm}

\def\eqref#1{equation~\ref{#1}}

\def\1{\bm{1}}

\DeclareMathAlphabet{\mathsfit}{\encodingdefault}{\sfdefault}{m}{sl}
\SetMathAlphabet{\mathsfit}{bold}{\encodingdefault}{\sfdefault}{bx}{n}

\usepackage{booktabs}  
\usepackage{multirow}  
\usepackage{siunitx}  
\usepackage{makecell}  
\usepackage{algorithm}
\usepackage{algpseudocode}
\usepackage{url}
\usepackage{graphicx}
\usepackage{listings}  
\usepackage{upquote}
\usepackage{xcolor}    
\usepackage{soul} 
\usepackage{pifont}
\usepackage{subcaption}
\usepackage{fvextra}
\usepackage[utf8]{inputenc}
\usepackage[export]{adjustbox} 
\usepackage{svg}
\usepackage{enumitem}
\usepackage{tabularx} 

\DeclareUnicodeCharacter{1EAD}{\d{\^a}}
\DeclareUnicodeCharacter{1EC5}{\~{\^e}}

\usepackage[most]{tcolorbox}

\newtcolorbox[list inside=prompt,auto counter,number within=section]{prompt}[1][]{
    colbacktitle=black!60,
    fonttitle=\small,
    coltitle=white,
    fontupper=\footnotesize,
    boxsep=3pt,
    left=0pt,
    right=0pt,
    top=0pt,
    bottom=0pt,
    boxrule=1pt,
    #1,
    breakable,              
}

\usepackage{nicematrix}
\definecolor{beige}{rgb}{0.9254901960784314, 0.9254901960784314, 0.9058823529411765}
\definecolor{purple}{rgb}{0.5098039215686274, 0.4235294117647059, 0.4980392156862745}

\definecolor{skyblue}{rgb}{0.5294117647058824, 0.807843137254902, 0.9215686274509803}   
\definecolor{steelblue}{rgb}{0.27450980392156865, 0.5098039215686274, 0.7058823529411765} 
\definecolor{royalblue}{rgb}{0.2549019607843137, 0.4117647058823529, 0.8823529411764706} 
\definecolor{mintgreen}{rgb}{0.596078431372549, 0.984313725490196, 0.596078431372549}   
\definecolor{seagreen}{rgb}{0.1803921568627451, 0.5450980392156862, 0.3411764705882353} 
\definecolor{tealgreen}{rgb}{0.0, 0.5019607843137255, 0.5019607843137255}              

\definecolor{c-navy}{rgb}{0,0.08,0.45}
\definecolor{c-blue}{RGB}{29,59,105}
\usepackage{hyperref}
\hypersetup{ %
colorlinks=true,
linkcolor=c-navy,
citecolor=c-navy,
filecolor=c-navy,
urlcolor=c-navy,
}
\usepackage{url}

\newcommand{\githublogo}[1]{\hspace{2pt}\includegraphics[width=13pt, valign=c]{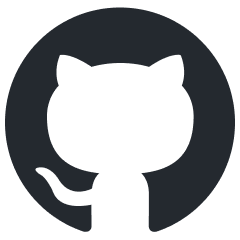}\hspace{2pt} \texttt{#1}}
\newcommand{\hflogo}[1]{\includegraphics[width=15pt, valign=c]{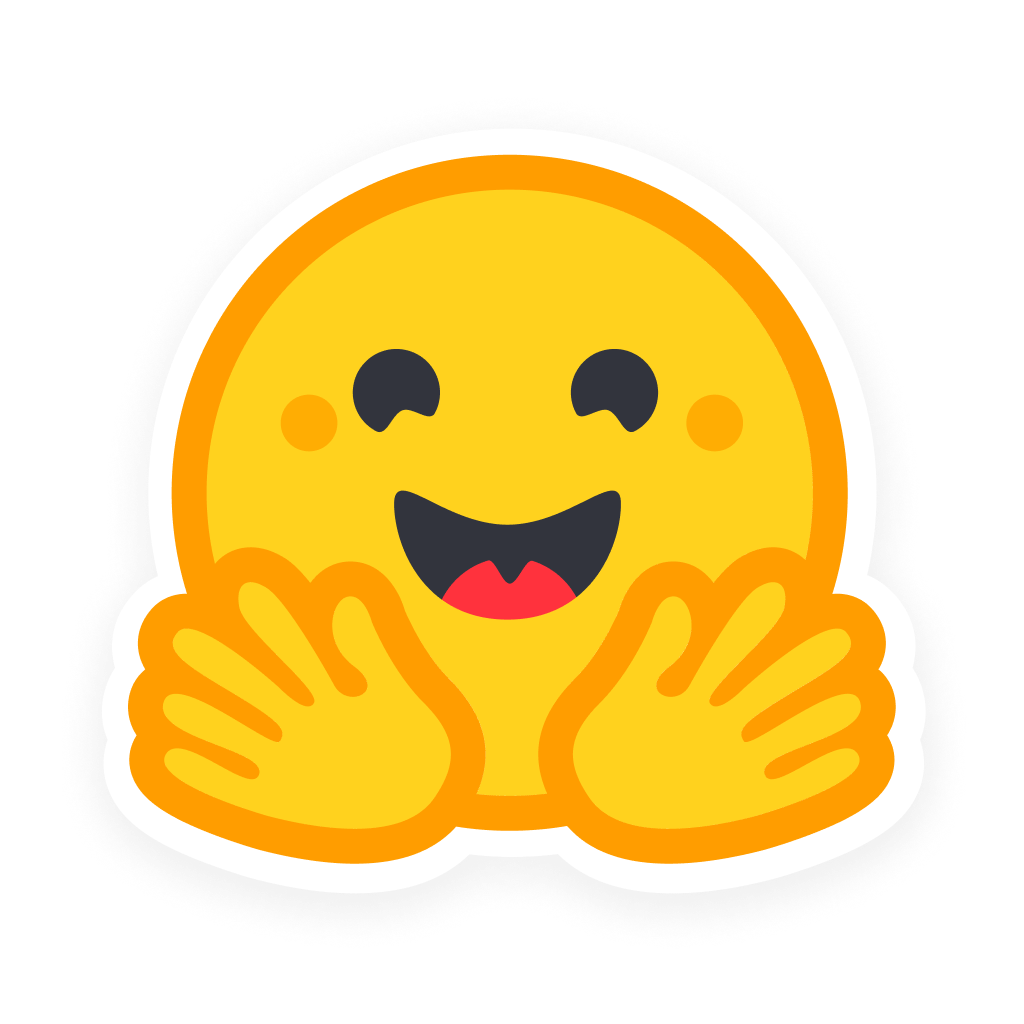} \texttt{#1}}
\newcommand{\datalogo}[1]{\hspace{1pt}\includegraphics[width=15pt, valign=c]{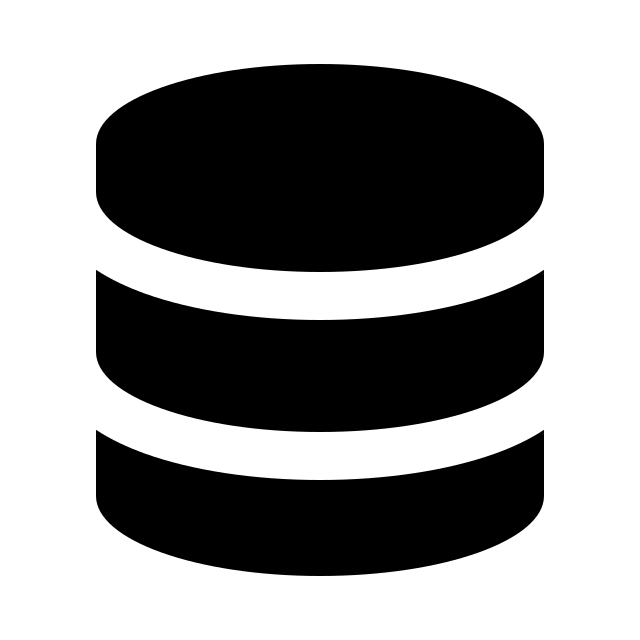}\hspace{1pt} \texttt{#1}}
\newcommand{\benchmarklogo}[1]{\hspace{1pt}\includegraphics[width=15pt, valign=c]{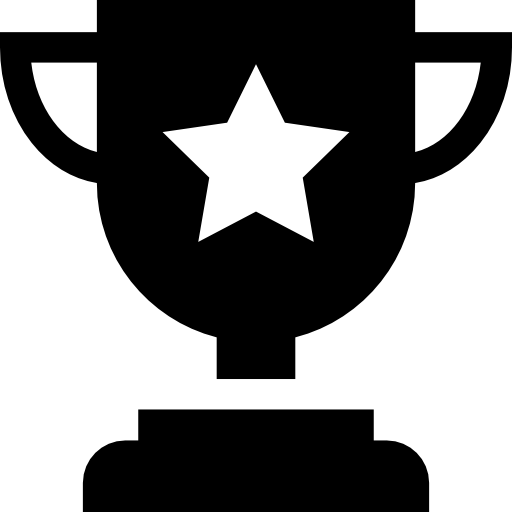}\hspace{1pt} \texttt{#1}}

\title{\papertitle \\[1em]
\centering
 \\[2mm]
\modellogos
}
\author{Project Polyglot\thanks{Polyglot is a project supported by Germany’s Federal Ministry of Education and Research (BMBF) and North Rhine-Westphalia’s Ministry of Culture and Science (MWK) under the TRA Sustainable Futures initiative at the University of Bonn.} \\ \\ 
\textbf{Leads:} 
Shiza Fatimah\textsuperscript{1,2},
Nicholas Kluge Corrêa\textsuperscript{1,2} \\ \\
\textbf{Core Team:}
Shiza Fatimah\textsuperscript{1,2},
Sophia Falk\textsuperscript{3},
Aniket Sen\textsuperscript{4},
Nicholas Kluge Corrêa\textsuperscript{1,2} \\ \\
\textbf{Other Contributors:} 
Lucie Flek\textsuperscript{1,2},
Florian Mai\textsuperscript{1,2} \\ \\
\textbf{Affiliations:}\\
  \textsuperscript{1}Bonn-Aachen International Center for Information Technology (b-it) / CAISA Lab\\
  \textsuperscript{2}Lamarr Institute for Machine Learning and Artificial Intelligence \\
  \textsuperscript{3}Bonn Sustainable AI Lab\\
  \textsuperscript{4}Helmholtz-Institut für Strahlen- und Kernphysik\\
}
\newcommand{\papertitle}{\textbf{Raising Bars, Not Parameters: LilMoo Compact Language Model for Hindi}}

\newcommand{\modellogos}{\resizebox{7cm}{!}{\includegraphics{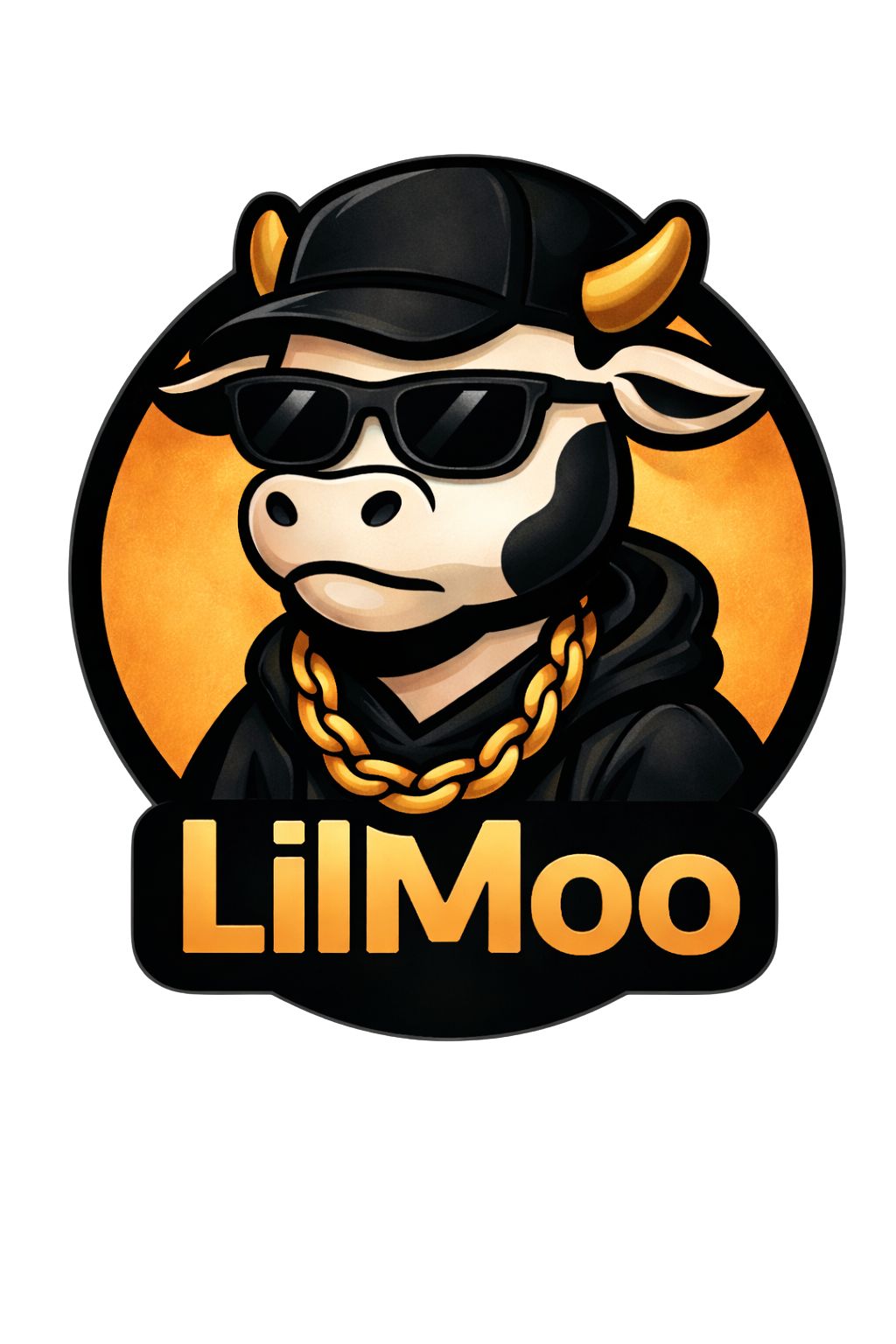}}}  %
\definecolor{lilmoocolor}{HTML}{F4B400}

\newcommand{\gigalekh}{\textcolor{lilmoocolor}{GigaLekh}\xspace}
\newcommand{\lilmoo}{\textcolor{lilmoocolor}{\textbf{LilMoo}}\xspace}
\newcommand{\lilmoohi}{\textcolor{lilmoocolor}{\textbf{LilMoo-v0.1}}\xspace}
\newcommand{\lilmoohien}{\textcolor{lilmoocolor}{\textbf{LilMoo-v0.2}}\xspace}

\definecolor{darkgreen2}{HTML}{009B55}
\definecolor{myblue2}{HTML}{29abe2}
\definecolor{myorange2}{HTML}{f7931e}

\definecolor{mypurple2}{HTML}{9823FF}

\begin{document}

\maketitle

\newpage

\begin{abstract}
The dominance of large multilingual foundation models has widened linguistic inequalities in Natural Language Processing (NLP), often leaving low-resource languages underrepresented. This paper introduces \lilmoo, a 0.6-billion-parameter Hindi language model trained entirely from scratch to address this gap. Unlike prior Hindi models that rely on continual pretraining from opaque multilingual foundations, \lilmoo is developed through a fully transparent and reproducible pipeline optimized for limited compute environments. We construct a high-quality Hindi corpus (\gigalekh) filtered through both heuristic and learned (LLM-as-a-judge) methods, complemented by bilingual augmentation with curated English data. Using this dataset, we explore various training recipes for small-scale language models. Across comprehensive evaluation suites, \lilmoo consistently outperforms comparably sized multilingual baselines such as Qwen2.5-0.5B and Qwen3-0.6B, demonstrating that well-designed language-specific pretraining can rival large multilingual models at the sub-billion-parameter range.
\end{abstract}

\newpage
\tableofcontents

\newpage
\section{Introduction}
\label{sec:introduction}

The Natural Language Processing (NLP) landscape has been reshaped by deep learning, particularly through self-supervised learning (SSL) frameworks that leverage vast unlabeled data to learn rich representations \citep{balestriero2023cookbook}. Transformer architectures, introduced in foundational works by \cite{vaswani2017attention}, enable parallelizable training on massive corpora and the development of models that excel in tasks from translation to natural language generation. This new SSL-based training paradigm contrasts with earlier supervised methods, offering generality and transferability across downstream applications.

Empirical scaling laws underscore the paradigm's reliance on compute. As shown time and time again, performance tends to increase as certain dimensions of neural training, such as data, model size, and compute, are scaled \citep{hoffmann2022training}. For instance, Qwen2.5 \citep{hui2024qwen2} released a series of dense decoder models ranging from 0.5B up to 72B parameters, all trained on a dataset of $\sim$18 trillion tokens. Qwen3 comprises six dense models, along with two Mixture-of-Experts (MoE) variants, with dense model sizes ranging from 0.6B to 32B parameters, and is trained on a much larger dataset ($\sim$36 trillion tokens) (\cite{yang2025qwen3}). Meanwhile, smaller open models are also trained at similar scales. AllenAI's OLMo2 series was trained on up to $\sim$5 trillion tokens \citep{olmo20242}, while the "Smol" series used $\sim$11 trillion tokens for their SmolLM2 (1.7B) and SmolLM3 (3B) models \citep{allal2025smollm2,bakouch2025smollm3}.

However, this compute- and data-intensive paradigm has inadvertently created a schism in the field, leading to a growing linguistic divide that marginalizes many languages. Of the roughly 7,000 languages spoken worldwide---most of which are low-resource and at risk of erosion---high-quality datasets exist for only a small minority \citep{cohere2024languagegap}. Furthermore, suppose we expand the notion of a "low-resource language" to encompass not only a limited digital footprint or sparse web data. In that case, it becomes evident that structural factors, such as inadequate infrastructure, limited technical expertise, and insufficient funding, also play a critical role in perpetuating this imbalance. Although the number of open-weight models, such as the Qwen and LLaMA families, continues to grow, these releases provide only partial insight into the development process behind these artifacts. Truly open-source and reproducible research at the frontier of foundation model development remains the domain of a limited number of institutions (e.g., the OLMo models \citep{groeneveld2024olmo,olmo20242} from AllenAI and the SmolLM series \citep{allal2025smollm2,bakouch2025smollm3} from Hugging Face).

In response to this linguistic disparity, the field has increasingly turned toward multilingual foundation models as a potential remedy. These models are designed to operate across dozens of languages, extending the benefits of large-scale language modeling beyond English and other high-resource languages. These models, when released with open weights (and permissible licensing), allow developers to extend the capabilities of these foundations, narrowing them to specific contexts through a process often referred to as continual pretraining \citep{liu2021continual, ke2023continual, zhou2024continual, joshi2025adapting, nag2025efficient}, i.e., further pretraining a foundation model on a specific domain (such as a language) of choice.

Regarding the multilingual approach, recent efforts exemplify this trend. The Qwen family \citep{hui2024qwen2,yang2025qwen3} variants are trained on extensive datasets covering over 30 languages, while Llama \citep{grattafiori2024llama} and Gemma \citep{team2024gemma} also emphasize multilingual and cross-domain performance. The same can be said for the Mistral models \citep{rastogi2025magistral}, the Falcon models \cite{zuo2025falcon}, IBM's Granite series \citep{granite2025}, and the more recent SmolLM3, developed by Hugging Face \citep{bakouch2025smollm3}, to name a few.

Despite their promise, multilingual foundation models remain constrained by data imbalance and representation asymmetries. High-resource languages tend to dominate the quantity and quality of training data, often overshadowing low-resource ones whose inclusion may be nominal rather than substantive. Consequently, monolingual models frequently outperform multilingual counterparts on tasks in their respective languages ( \citep{virtanen2019multilingual, martin2020camembert, armengol2021multilingual, correa2025tucano}. Thus, multilingual models offer breadth but often lack depth in individual languages.

Similarly, while the Continual Pretraining (CPT) approach offers a practical path to rapidly producing proficient models under significant resource constraints, it also presents important limitations --- particularly when applied to black-box foundation models whose pretraining data, recipes, and hyperparameters remain undisclosed. In such cases, although CPT enables efficient adaptation to new domains or languages, it yields limited insight into the underlying learning dynamics, making it challenging to attribute performance gains to specific factors. Nonetheless, the field's current state still favors models fine-tuned on large, opaque foundations — such as Qwen or LLaMA — whose original pretraining details are often proprietary. This lack of transparency constrains scientific understanding of what drives performance and limits reproducibility. Consequently, while CPT remains a valuable strategy for extending existing models to new languages or domains, it is not a substitute for open, transparent, and reproducible pretraining practices that can more fundamentally advance linguistic inclusivity and model interpretability.

This brings us to our work, which focuses on Hindi, a widely spoken language that remains underrepresented in the NLP landscape. As of 2025, Hindi ranks third among the most widely spoken languages globally \citep{ethnologue200}. Nevertheless, it remains classified as a low-resource language, primarily due to the scarcity of publicly accessible digital data and insufficient research investment. An additional challenge for Indic languages such as Hindi is the pervasive integration of English, stemming from colonial history, the entrenched use of English in education and media, and its functional role in digital communication \cite{sengupta2024social, parshad2014india, srivastava2020understanding}. Consequently, much of the text data available online for pretraining models contains English contamination to varying extents.

We introduce \lilmoo, a 0.6B-parameter Hindi language model trained from scratch on a curated dataset of high-quality Hindi text. This marks a deliberate step toward building native foundations that prioritize depth over diluted multilingual breadth. Our work explores how augmenting the training data with high-quality English text can enhance model performance, leveraging the symbiotic relationship between Indic languages and English in bilingual contexts. By strategically incorporating high-quality English corpora, we demonstrate gains in cross-lingual transfer and overall robustness without compromising the model's core proficiency in its native language.

Regarding assets, this work also presents a fully open and reproducible recipe for training language models in low-resource settings. Building on other open research initiatives, we adapt and modify existing techniques and recipes to fit our needs and constraints, including limited compute budgets and sparse high-quality data. The result is a series of artifacts designed to advance NLP research in Hindi, including high-quality datasets, evaluation benchmarks, models, and source code.

All code, datasets, and models are released under permissive licenses to facilitate further research and development in Hindi NLP.

\textbf{Our main contributions are:}

\begin{itemize}
    \item \textbf{Large-scale Hindi corpus.} A high-quality Hindi text dataset, filtered using heuristic and learning-based methods. The Hindi dataset (\gigalekh) contains approximately 90 billion tokens from diverse domains, with a knowledge cut-off of December 2025. Auxiliary annotated datasets for training quality and toxicity filters tailored to the language, together with two lightweight models for these tasks, enable scalable data cleaning without external dependencies.
    \item \textbf{Annotation and filtering models.} Classifiers for education quality and toxicity annotation trained on novel auxiliary datasets and released for use by the research community.
    \item \textbf{Base models.} With two versions: a native-only model trained entirely on Hindi (\lilmoohi) and a mixed-language version trained with high-quality English text (\lilmoohien).
    \item \textbf{Hindi evaluation harness.} An implementation of a comprehensive and reproducible evaluation suite for benchmarking NLP models on Hindi tasks.
    \item \textbf{Fully open release.} To promote reproducibility, all datasets, models, training recipes, and evaluation code are openly released under permissive Apache 2.0 licenses.
\end{itemize}

This work is part of the broader Polyglot project. The methodology and findings presented here extend to additional language-specific studies conducted within the same framework, including Portuguese (e.g., the \href{https://huggingface.co/collections/Polygl0t/tucano2}{Tucano 2} series) and Bengali (e.g., \href{https://huggingface.co/collections/Polygl0t/liltii}{LilTii}). For further details on these parallel efforts and associated resources, please refer to the Polyglot project page: \href{https://huggingface.co/Polygl0t}{huggingface.co/Polygl0t}.

\newpage
\section*{Released Assets}
\textbf{Monolingual Models}
\\[1mm]
\begin{small}
\hflogo{\href{https://huggingface.co/Polygl0t/LilMoo-v0.1}{Polygl0t/LilMoo-v0.1}} {(Hindi)}\\
\end{small}

\textbf{Bilingual Models}
\\[1mm]
\begin{small}
\hflogo{\href{https://huggingface.co/Polygl0t/LilMoo-v0.2}{Polygl0t/LilMoo-v0.2}} {(Hindi+English)}\\
\end{small}

\textbf{Datasets}
\\[1mm]
\begin{small}
\datalogo{\href{https://huggingface.co/datasets/Polygl0t/gigalekh-v1}{Polygl0t/gigalekh-v1}}\\
\end{small}

\textbf{Evaluation Harness}
\\[1mm]
\begin{small}
\benchmarklogo{\href{https://github.com/Polygl0t/lm-evaluation-harness/tree/polyglot_harness_hindi}{Polygl0t/lm-evaluation-harness}} { (Hindi branch)}\\
\end{small}

\textbf{Code}
\\[1mm]
\begin{small}
\githublogo{\href{https://github.com/Polygl0t/llm-foundry}{Polygl0t/llm-foundry}} \\
\end{small}

\newpage

\newpage
\section{Related Work}
\label{sec:related_work}

The development of language models for Indic languages, including Hindi, has gained momentum in recent years, driven by the broader shift toward multilingual and open-source NLP. However, progress remains fragmented, with most efforts relying on continual pretraining of existing multilingual foundations rather than on creating a complete training pipeline, and often grappling with data quality, reproducibility, and comprehensive evaluation issues. This section reviews key contributions in Hindi language modeling, highlighting their methodologies, deliverables, and limitations, before discussing how our work, \lilmoo, addresses persistent gaps

\subsection{Hindi Language Models}

Research on Indic language models has predominantly focused on multilingual approaches, with comparatively limited work dedicated exclusively to Hindi, primarily due to data scarcity. To address this limitation, several models have been trained on multiple Indic languages simultaneously. Notable examples include IndicBERT \citep{kakwani2020indicnlpsuite} and MuRIL \citep{khanuja2021muril}, which are multilingual ALBERT (\cite{lan2019albert}) and BERT \citep{devlin2018bert} based encoder models, respectively. IndicBERT is pretrained on 8.8 billion tokens across 11 Indian languages, and later fine-tuned on the IndicGLUE \citep{kakwani2020indicnlpsuite} benchmark tasks. MuRIL is a 236-million-parameter multilingual BERT-base encoder model supporting 17 Indic languages.

Despite the dominance of multilingual models, several efforts have been made to develop Hindi-specific models. A common trend among these is fine-tuning large-scale foundation models for Hindi instruction-following tasks. For instance, LLaMA3-Gaja-Hindi-8B is a bilingual Hindi–English model, fine-tuned from LLaMA-3 (8B parameters) on a curated dataset of translated instructional pairs. Apart from the model release \citep{llama3_gaja_hindi}, no additional information, including its training dataset, training recipe, or any associated research publication, has been publicly disclosed at the time of our writing.

Another line of research extends large multilingual models through continued pretraining on Hindi text. One example is Nemotron-4-Mini-Hindi \citep{joshi2025adapting}, a 4 billion parameter bilingual Hindi–English model based on Nemotron-Mini-4B \citep{muralidharan2024compact}. It underwent continued pretraining on 400 billion tokens, with an equal split between 200 billion Hindi and 200 billion English tokens, adapted from the primarily English-trained Nemotron-4B. Another model that underwent continued pre-training is LLaMA-3-Nanda-10B \cite{choudhury2025llama}. It's a 10B-parameter bilingual Hindi-English decoder-only model based on LLaMA-3. It underwent continued pretraining on 65B Hindi tokens using block expansion, followed by instruction-tuning. 

A smaller subset of work focuses on models trained entirely from scratch on curated Hindi corpora. Key contributions include HindiLLM introduced by \cite{chouhan2025hindillm}, a GPT-2-based model released in two variants with 124M and 354M parameters. HindiLLM was developed through unsupervised pretraining on a 37.3 GB Hindi corpus followed by supervised fine-tuning. Both model variants are released on Hugging Face; however, as of the time of writing, the source code used to develop this model remains unavailable. Another available Hindi model is OpenHathi \citep{openhathi}, developed by SarvamAI \footnote{\href{https://www.sarvam.ai/}{www.sarvam.ai}}. According to its model card description, it is a 7B-parameter model based on LLaMA-2 \citep{touvron2023llama2}, trained on Hindi, English, and Hinglish text. Beyond this brief description, no further technical documentation or training details are available.

\subsection{Challenges and Motivation}

Across these works, common threads emerge:

\begin{itemize}
\item Due to computational and data constraints, most approaches rely on continual pretraining or fine-tuning of existing foundation models whose original pretraining data is not disclosed.
\item Another notable trend is that many models are either trained across multiple Indic languages or incorporate substantial amounts of English text, whereas only a small number of works develop native models from scratch.
\item While several works have incrementally nurtured the Hindi NLP community with baselines, datasets, and evaluation harnesses, there are still gaps in the reproducibility of a complete development stack for LLM development, especially regarding transparency in source code, dataset, and documentation on how to use it.
\end{itemize}

Fortunately, several recent open research efforts provide a foundation for addressing these gaps. Instances such as OlMO \citep{olmo20242}, SmolLM \citep{allal2025smollm2}, LLM360 \citep{liu2023llm360}, DeepSeek \citep{bi2401deepseek}, and Apertus \citep{apertus2025apertus} offer valuable insights and methodologies for efficient pretraining and model development. However, these training recipes often assume access to large compute resources and extensive datasets, which may not align with the constraints faced in low-resource language contexts. Therefore, while these works provide a helpful starting point, adapting their approaches to the specific needs and limitations of low-resource languages such as Hindi remains a critical challenge, which we attempt to address in this work.

\newpage
\section{Pretraining Data}
\label{sec:pretraining_corpus}

When working with low-resource languages, scarce data is among the most challenging aspects. While large-scale web crawls like CommonCrawl (CC), OSCAR \citep{abadji2022towards}, and many other derivatives from the CC initiative \citep{nguyen2024culturax,penedo2025fineweb2, burchell2025expanded, weber2411redpajama} provide a starting point, their coverage of low-resource languages is often limited. Moreover, these datasets often contain significant noise, including low-quality text, duplicates, and offensive content. To address these challenges, we adopted a multi-faceted approach to curate a high-quality Hindi corpus.

We present \gigalekh, a high-quality Hindi corpus. \gigalekh contains approximately 90 billion tokens across 83 million documents. To ensure content quality, we leverage Qwen2.5-32B \citep{hui2024qwen2} as an LLM judge to annotate documents for educational relevance and toxicity. These annotations are then distilled into lightweight classifiers, enabling corpus-scale filtering. The resulting corpus includes rich metadata, such as educational and toxicity scores. Our pipeline filtering pipeline is based on the work done by FineWeb2 \citep{penedo2025fineweb2}.

\subsection{Corpus Construction Pipeline}

Our dataset is assembled by combining curated textual resources with large-scale web data. We begin with publicly available datasets hosted on Hugging Face, which typically contain partially processed text and therefore require minimal preliminary cleaning. To broaden topical diversity and ensure inclusion of recent material, we supplement these sources with newly released Common Crawl snapshots. All collected data are then processed through a unified multi-stage pipeline designed to standardize quality and format across heterogeneous inputs. The overall architecture follows the general principles of the FineWeb2 framework \citep{penedo2025fineweb2}, adapted for Hindi.

\paragraph{Stage 1: Source Screening and Text Extraction}
Documents originating from web archives are first filtered using a URL blocklist to exclude undesirable or low-quality domains. Textual content is then extracted from WARC files using Trafilatura \citep{barbaresi2021trafilatura}. This step is omitted for Hugging Face datasets because their contents are already pre-cleaned text.

\paragraph{Stage 2: Language Verification}
Language identification is performed sequentially using two models. The first pass employs the FT176 classifier \citep{joulin2016fasttext}, retaining documents predicted as Hindi with confidence scores of at least 0.69.This fast, coarse-grained model allows
broad coverage and ensures that scarce Hindi data is not discarded prematurely. Documents that pass this stage are re-evaluated using GlotLID \citep{kargaran2023glotlid}; only those meeting the same confidence requirement are preserved. The combination of a fast, broad filter followed by a more precise model improves overall reliability.

\paragraph{Stage 3: Heuristic Content Filtering}
To operationalize document quality, we apply rule-based filters that capture measurable indicators of linguistic well-formedness and structural coherence. These filters remove texts containing excessive repetition, malformed sentences, abnormal word-length distributions, missing punctuation, or insufficient stopword presence. Such signals commonly indicate templated pages, scraping artifacts, or automatically generated content. Thresholds and heuristics are tuned to reflect properties of Hindi text.

\paragraph{Stage 5: Duplicate Removal}
To reduce redundancy, we apply similarity-based deduplication using MinHash \citep{broder1997resemblance}. Each document is converted into 5-gram shingles, hashed into compact signatures, and compared via locality-sensitive hashing. We use 14 buckets with 8 hashes per bucket and \texttt{xxHash} for efficient computation. This approach scales well to large corpora while reliably identifying near-duplicate documents.

\paragraph{Stage 6: Learned Quality Filtering}
Finally, documents that pass all previous stages are evaluated using a learned filtering component that assigns quality metadata such as educational quality and toxicity using an LLM-as-a-Judge filtering approach. Because this stage involves a distinct modeling framework and training procedure, it is described separately in Section~\ref{sec:llmfilter}.

\subsection{Learned Quality Filtering}
\label{sec:llmfilter}

Rule-based preprocessing improves structural consistency but cannot fully capture semantic properties such as educational relevance or contextual harmfulness. To incorporate these higher-level signals, we introduce a learned evaluation component that produces fine-grained annotations for documents passing earlier pipeline stages.

\paragraph{LLM Annotation}
We employ Qwen2.5-32B-Instruct \citep{hui2024qwen2} as an evaluator that assigns structured scores to sampled documents. This model was selected due to its multilingual capability, strong empirical performance, and permissive Apache-2.0 license, which allows redistribution of derived annotations. Each sampled document is assessed along two independent dimensions:

\begin{itemize}
\item educational usefulness on a five-level scale
\item toxicity on a five-level scale
\end{itemize}

The prompts used for scoring are provided in Appendix~\ref{appendix:llm-judge-prompts}. We run the evaluation on a random sample of 320{,}000 documents drawn from all parts of the dataset. Inference is performed using vLLM \citep{kwon2023efficient} on four NVIDIA A100-80GB GPUs. The labeled datasets are released publicly.

\paragraph{Training Classifiers}
Running a large model on the full corpus would be computationally expensive. Instead, we train smaller classifiers using the labeled data. We test IndicBERT \citep{kakwani2020indicnlpsuite} and Hindi-RoBERTa \citep{joshi2022l3cube} as high-performing encoder-based models to serve as a frozen feature extractor, enabling the classification head to be trained with minimal compute. Training details, hyperparameters, and full results for all classifiers are provided in Appendix~\ref{appendix:classifier-training}. Among the evaluated models, Hindi-RoBERTa achieved the strongest overall performance and was therefore selected for final filtering.

\subsection{Final Dataset Composition}

Once the full filtering pipeline was completed—including text extraction, language identification, heuristic quality checks, deduplication, and learned filtering with our trained classifiers—we applied two additional steps: (1) removing documents shorter than 50 tokens, and (2) separating documents with a toxicity score greater than 3. Instead of discarding these toxic documents, we retained them in a distinct subset, which can support future studies on toxicity detection and mitigation in NLP. The resulting Hindi dataset is named \gigalekh.\footnote{\href{https://huggingface.co/datasets/Polygl0t/gigalekh-v1}{Polygl0t/gigalekh-v1}}. A detailed breakdown of the dataset, including the contributions from each source, is available in Appendix~\ref{appendix:base-data-sources-statistics}.

Table~\ref{tab:gigalekh-stats} summarizes the size and token counts for the two subsets of \gigalekh. The \texttt{default} subset comprises the primary curated corpus used for pretraining, whereas the \texttt{excluded} subset contains documents filtered out due to high toxicity scores (documents with \texttt{toxic\_int\_score} $\geq$ 4). 

\begin{table}[h]
\centering
\caption{Statistics for subsets.}
\label{tab:gigalekh-stats}
\begin{tabular}{@{}lrrrr@{}}
\toprule
\textbf{Subset} & \textbf{Documents} & \textbf{Size} & \textbf{Tokens} \\
\midrule
\texttt{default} & 83,081,507 & 260.68 GB & 90,705,245,239 \\
\texttt{excluded} & 498,892 & 1.76 GB & 1,545,290,236 \\
\midrule
\textbf{Total} & 83,580,399 & 262.44 GB & 92,250,535,475 \\
\bottomrule
\end{tabular}
\end{table}

\newpage
\section{Tokenizer Design and Evaluation}
\label{sec:tokenization}

Efficient tokenization is crucial for training language models, as it determines how compactly text can be represented and directly affects training time, inference speed, and compute requirements \citep{finardi2021berta, larcher2023cabrita, correa2024teenytinyllama}. To effectively train our language model, we needed a tokenizer that could handle not only Hindi but also English and code data, reflecting the multilingual and multi-domain nature of our training corpus.

\subsection{Training Data}

We built the tokenizer using high-quality text from three sources: Hindi documents with Edu-score $\geq 3$, English content from the FineWeb-Edu dataset \citep{penedo2024fineweb}, and code samples from the Starcoder dataset \citep{li2023starcodersourceyou}, spanning 36 programming languages. This resulted in a training mixture of roughly 40\% Hindi, 40\% English, and 20\% code samples, with a total of 4,975,000 text samples. 

The tokenizer was trained from scratch using the SentencePiece library \citep{kudo2018sentencepiece}. To balance model expressiveness and GPU efficiency, we set the vocabulary size to 49,152 tokens. This choice keeps the embedding and output projection layers aligned with GPU-friendly dimensions (multiples of 128 and 256), thereby improving memory access patterns and computational throughput without unnecessarily increasing model size for a mono- or bilingual language model.

\subsection{Evaluation Metrics}

To assess tokenizer performance, we relied on two standard metrics:

\begin{itemize}
    \item \textbf{Subword Fertility (SF):} the average number of tokens per word. Lower values indicate more compact representations.
    \item \textbf{Proportion of Continued Words (PCW):} the fraction of words that are split into two or more tokens. Lower values suggest less aggressive splitting.
\end{itemize}

\subsection{Results and Comparison}

We evaluated the trained tokenizer on a test set of 31,500 words, including 15,000 Hindi words, 15,000 English words, and 1,500 code-mixed words. Table~\ref{tab:tokenizer-eval-hi} presents a comparison with several recent multilingual and Hindi-adapted tokenizers.

\begin{table}[h]
\centering
\caption{Tokenizer Evaluation Metrics Comparison Across Multiple Tokenizers for Hindi (Ordered by Fertility)}
\label{tab:tokenizer-eval-hi}
\begin{tabular}{@{}l c c c c c@{}}
\toprule
\textbf{Tokenizer} & \textbf{Tokens (31500 words)} & \textbf{Vocab Size} & \textbf{Fertility} & \textbf{PCW} & \textbf{UNK} \\
\midrule
\textbf{\lilmoo} & 46129 & 49152 & \textbf{1.46} & 0.50 & 0 \\
HindRoBERTa & 50697 & 250002 & 1.61 & 0.31 & 0 \\
Nemotron-Hindi-4B & 52111 & 256000 & 1.65 & 0.30 & 0 \\
IndicBERT & 54213 & 200000 & 1.72 & 0.40 & 2 \\
Gemma-2-2b & 54379 & 256000 & 1.73 & 0.47 & 0 \\
Llama-3.2 & 63891 & 128256 & 2.02 & 0.61 & 0 \\
HindiLLM & 78387 & 50007 & 2.49 & 0.66 & 0 \\
Qwen2.5 & 96634 & 151665 & 3.07 & 0.64 & 0 \\
\bottomrule
\end{tabular}
\end{table}

Our tokenizer, \lilmoo, achieves strong efficiency with a token count of 46,129 and a fertility of 1.46, outperforming most baselines, including multilingual models such as Qwen2.5 and Llama-3.2, which produce higher token counts and fertility values.

\newpage
\section{Infrastructure and optimizations}
\label{sec:infrastructure-optimization}

\subsection{Marvin}

Our training experiments were conducted on the Marvin HPC cluster \footnote{\href{https://www.hpc.uni-bonn.de/en/systems/marvin}{www.hpc.uni-bonn.de/en/systems/marvin}} at the University of Bonn. Marvin is a hybrid HPC system that features GPU partitions with NVIDIA A100-SXM4-80GB and NVIDIA A40-48GB, high-speed NVLink and InfiniBand NDR interconnects, a Lustre parallel file system for multi-petabyte storage, and a SLURM workload manager for job scheduling. All of our training experiments were performed using the nodes equipped with NVIDIA A100 GPUs. Concurrently, evaluations were conducted on the A40 nodes. Data preprocessing, filtering, and tokenization were performed on the cluster's CPU nodes. For our training runs, we used 2 nodes, each with 4 A100 GPUs, for a total of 8 GPUs/replicas in the distributed training setup. For evaluations, given that each A40 node is equipped with 8 GPUs, we set up our evaluation harness to evaluate different model checkpoints in parallel across multiple GPUs.

\subsection{Software Stack and Training Optimizations}

Although our model is relatively small (around 0.6B parameters), training still demands careful management of GPU memory and computational efficiency, especially when using larger batch sizes to improve convergence and generalization \citep{shallue2019measuring}. To meet these requirements, we designed a training pipeline that combines widely adopted software libraries with several performance-focused techniques.  

We built our stack on PyTorch \citep{paszke2019pytorch} and Hugging Face Transformers \citep{wolf-etal-2020-transformers}, running on Python 3.12.3 with CUDA 12.6.0. Because the model fits comfortably within a single A100 GPU, we opted for PyTorch’s native Distributed Data Parallel (DDP) rather than more complex frameworks such as Fully Sharded Data Parallel (FSDP) \citep{xu2020automatic, rajbhandari2020zero} or Megatron-LM \citep{shoeybi2019megatron}. In DDP, each GPU maintains a copy of the full model and optimizer, computes gradients independently on its batch, and synchronizes gradients via NCCL AllReduce during the backward pass. This approach simplifies implementation and provides near-linear scaling as GPUs are added, but it does require careful memory management when using large batches.  

To maximize efficiency and reduce memory overhead, we applied a series of optimizations. First, mixed-precision training with bfloat16 (BF16) was enabled, allowing faster computation while keeping the dynamic range of FP32. NVIDIA A100 tensor cores further accelerate matrix multiplications through TF32 mode, providing high throughput without sacrificing numerical stability. Second, we implemented Grouped-Query Attention (GQA) \citep{ainslie2023gqa} to reduce the memory footprint of multi-head attention: queries remain separate per head while keys and values are shared within each group. Third, activation checkpointing \citep{chen2016training} allows longer sequences and larger batches by storing only partial forward activations and recomputing them during backpropagation. Finally, we leveraged FlashAttention-2 \citep{dao2022flashattention} for memory-efficient attention computations and the Liger Triton kernel library \citep{hsu2025ligerkernel} for fused operations such as RMSNorm, RoPE, and SwiGLU.

Collectively, these optimizations enabled us to train the model efficiently on our infrastructure. To quantify hardware utilization, we measured the Model FLOP Utilization (MFU) \citep{chowdhery2023palm}, which estimates how effectively the model uses GPU compute resources. Our training runs achieved approximately 70\% MFU on the A100 GPUs, reflecting a balance between throughput, memory usage, and computational efficiency.

\newpage
\section{Pretraining}
\label{sec:pretraining}

We train two 670-million-parameter models based on the original Llama architecture. Both models are trained on approximately the same number of tokens. The first version (\lilmoohi), is trained exclusively on Hindi using a single-stage cosine learning rate decay schedule. In contrast, the second version (\lilmoohien) is trained on a mixture of Hindi and English data, in a multi-stage training recipe. Further details regarding training configurations and data composition are provided in the following sections.

\subsection{Model Architecture}

The Llama architecture builds upon the original GPT-style transformer \citep{radford2018improving} and incorporates several enhancements that have been shown to improve performance and training stability, like the use of RMSnorm \citep{zhang2019root} for normalization, RoPE positional embeddings \citep{zhang2019root}, and SwiGLU activations \citep{shazeer2020glu}. As an easy-to-use reference implementation of this architecture, we used the \texttt{transformers} library implementation of Llama \citep{wolf-etal-2020-transformers}, which is equivalent to implementations like SmoLLM2 \citep{allal2025smollm2} and TinyLlama \citep{zhang2024tinyllama}. The dimensions of our models are summarized in Table \ref{tab:architecture}.

\begin{table}[h]
\centering
\caption{Architectural Specifications.}
\label{tab:architecture}
\begin{tabular}{@{}l c@{}}
\toprule
\textbf{Parameter} & \textbf{Value} \\
\midrule
Activation & SwiGLU \\
Hidden$_d$ (Residual Stream) & 1,536 \\
Feed-forward (MLP) & 3,072 \\
Context Length & 4,096 tokens \\
Attention Heads & 16 \\
Attention Head$_d$ & 96 \\
Layers & 28 \\
Key/Value Heads & 8 \\
Tied Embeddings & True \\
Vocabulary Size & 49,152 \\
\bottomrule
\end{tabular}
\end{table}

Our choice of model architecture and size was guided by three primary considerations: GPU usage efficiency, model expressiveness, and training saturation behavior. These factors jointly informed both the structural design of the transformer and the overall parameter scale.

\begin{itemize}
    \item \textbf{GPU Usage Efficiency:} From a hardware perspective, we followed established heuristics\footnote{\url{https://docs.nvidia.com/deeplearning/performance/index.html}} to align key architectural dimensions—such as hidden size, intermediate size, and vocabulary size—to powers of two and GPU-friendly divisibility constraints. This alignment improves tensor core utilization (particularly in TF32 mode) and reduces performance degradation caused by inefficient tiling or wave quantization.

    \item \textbf{Model Expressiveness:} We adopted a \textit{``deep and slim''} configuration, increasing depth while keeping width moderate. Recent studies (e.g., MobileLLM \citep{liu2024mobilellm}, SmolLM2 and SmolLM3 \citep{allal2025smollm2,bakouch2025smollm3}, ModernBERT \citep{warner2025smarter}) suggest that deeper architectures at small-to-mid scales improve generalization and reasoning capabilities compared to wider but shallower models.

    \item \textbf{Saturation and Scaling Behavior:} The overall parameter count was kept at a moderate scale based on insights from Pythia \citep{biderman2023pythia} and LLaMA \citep{touvron2023llama2}, which indicate that models above a certain size threshold (approximately 0.4B parameters) continue to benefit from extended training without early saturation, i.e., without rapid diminishing returns in performance improvements.
\end{itemize}

Meanwhile, it is worth noting that this model size (i.e., $\sim$0.6B parameters) allows us to perform experiments more efficiently, given the compute resources at our disposal, and to make a fair comparison against the Qwen baselines, which have similar parameter counts.

\subsection{Training Recipes}

Our primary research objective is to evaluate whether language-specific pretraining can achieve performance comparable to that of large-scale multilingual pretraining while requiring substantially less computational resources, at the sub-billion parameter scale. To investigate this, we designed two training recipes and compared their results with previously established multilingual baselines (i.e., Qwen2.5 and Qwen3). These recipes differ in data mixtures, training schedules, and overall strategies, allowing us to explore the trade-offs between monolingual depth and bilingual augmentation.

\begin{itemize}

    \item \textbf{Simple/Single-Stage (v0.1):} We used only native-language (Hindi) text with a cosine learning rate decay and warmup for this recipe. This approach provides a simple monolingual baseline that explores the performance achievable with a vanilla training setup. The model trained with this recipe is called \lilmoohi.
    
    \item \textbf{Multi-Stage Recipe (v0.2):} We employed a multi-stage training approach with a mixture of Hindi and English text for this recipe. The stages progressively refine the data mixture, starting with a balanced mix of Hindi and English educational content and later up-sampling higher-quality subsets. This approach is inspired by recipes such as those explored in the development of OlMo2 \citep{olmo20242} and the SmolLM series \citep{allal2025smollm2,bakouch2025smollm3}. The model trained with this recipe is \lilmoohien.
    
\end{itemize}

\subsection{Common Training Configuration}

Both recipes share a set of core parameters to ensure consistency across our comparisons. For example, both recipes use a total batch size of 2,097,152 tokens, a micro-batch size of 262,144 (16 x 4 gradient accumulation steps), and BF16 precision with TF32 enabled to enable efficient training on NVIDIA A100 GPUs. Both recipes use the AdamW optimizer with the same hyperparameter settings (maximum learning rate of 7e-4, $\beta_{1}$=0.9, $\beta_{2}$=0.95, $\epsilon$=1e-8) (see Table~\ref{tab:shared-training-params}).

\begin{table}[h]
\centering
\caption{Shared Training Hyperparameters for Both Recipes}
\label{tab:shared-training-params}
\begin{tabular}{@{}l l@{}}
\toprule
\textbf{Parameter} & \textbf{Value / Description} \\
\midrule
Total batch size & 2,097,152 tokens \\
Precision & BF16 with TF32 enabled \\
Optimizer & AdamW \\
Maximum learning rate & 7e-4 \\
Adam $\beta_1, \beta_2$ & 0.9, 0.95 \\
Adam $\epsilon$ & 1e-8 \\
Weight decay & 0.1 applied selectively \\
Excluded from weight decay & Biases, LayerNorm, Embeddings \\
Gradient Clipping & Maximum norm = 1.0 \\
Checkpointing Frequency & Every 5,000 steps (10 billion tokens) \\
\bottomrule
\end{tabular}
\end{table}

To further enhance training stability, we selectively apply weight decay. Specifically, we split the model parameters into two groups: one with weight decay (0.1) applied and one without. Parameters such as biases, layer normalization weights, and embedding tokens are excluded from weight decay. The choice to exclude embedding layers from weight decay is inspired by findings from OLMo 2 \citep{olmo20242} and SmolLM 3 \citep{bakouch2025smollm3}, which demonstrated that this modification leads to more stable training dynamics. Furthermore, both recipes incorporate gradient clipping with a maximum norm of 1.0 to prevent exploding gradients during training. Both recipes evaluate the model on a held-out validation set of 20,000 sequences, with validation performed every 5,000 steps. This validation set is randomly sampled from all the data sources for each recipe, ensuring the model's performance is monitored consistently throughout training. We checkpoint the model at each validation step for further evaluation on our harness.

To determine the common batch size (i.e., 2,097,152 tokens) and maximum learning rate (i.e., $7e-4$) shared among recipes, we relied on the heuristics established in the DeepSeek LLM paper \citep{bi2401deepseek}. Specifically, we adopted the empirical scaling laws proposed in the DeepSeek LLM paper, which relate the compute budget ($C$) to optimal hyperparameters. In these laws, both batch size and learning rate follow predictable power-law relationships concerning total compute. To estimate our compute budget ($C$), we applied the adjusted DeepSeek formulation---a refinement of the standard PaLM expression ($C = 6ND$)---that accounts for non-embedding FLOPs per token:
\[
C =
\left(
72 n_{\text{layer}} d_{\text{model}}^{2}
+
12 n_{\text{layer}} d_{\text{model}} \ell_{\text{seq}}
\right) D
\]

By plugging in our model's dimensions and the estimated dataset sizes for our two training recipes, we obtained the compute budget $C$ for our experiments. Using the DeepSeek scaling heuristics:

\[
\text{Max Learning Rate} = 0.3118 \cdot C^{-0.125}
\]

\[
\text{Batch Size} = 0.2920 \cdot C^{0.3271}
\]

We then derived the corresponding optimal hyperparameters for a common batch size of approximately 2 million tokens and a maximum learning rate of $7e-4$. The batch size was rounded to the nearest power of two ($2^{21}$) for hardware efficiency.

\subsection{Recipe 1: Simple (Single-Stage) Recipe}

Our primary research hypothesis is that language-specific pretraining can achieve performance comparable to large-scale multilingual pretraining, especially when the target language benefits from cross-lingual synergies with another language \citep{kakwani2020indicnlpsuite,ogueji2021small, chang2024multilinguality}. We first designed a "simple" or single-stage training recipe to test this hypothesis. This recipe focuses exclusively on native-language(Hindi) text to establish a monolingual baseline. A baseline that would allow us to assess whether adding English text to a multi-stage recipe provides any benefits.

Given the relative scarcity of high-quality text in Hindi compared to English, our simple recipe emphasizes depth and the limits of what can be achieved in a data-constrained setting. Following the insights from \cite{muennighoff2023scaling}, we repeated our core dataset for 5 epochs, yielding a total training volume of approximately $\sim$375 billion tokens. For this training recipe, the data mixture was not organized into stages or a curriculum; instead, the model was trained from scratch on its entire language-specific dataset, with the dataset shuffled at the beginning of each epoch. We characterize this approach as "simple" because it employs a straightforward training strategy, as documented in the development of the Pythia suite \citep{biderman2023pythia} and in earlier LLMs (e.g., GPT 2).

Moreover, this recipe uses a cosine learning rate decay scheduler with an initial warmup phase, a common choice for stable convergence in transformer pretraining. The learning rate starts at zero and increases linearly to a peak of $7e-4$ over the first 1,000 optimization steps. After the warmup phase, the learning rate decays following a cosine schedule down to a minimum learning rate of $7e-5$, which is held for 10\% of the total training steps. In total, according to the $C = 6ND$ formulation from \cite{chowdhery2023palm}, this recipe requires approximately $\sim 1.35 \times 10^{21}$ FLOPs to complete training.

\subsection{Recipe 2: Multi-Stage Recipe}

The second training recipe explores a more complex multi-stage approach that incorporates native-language (Hindi) and high-quality English text. As sources of high-quality English text, we selected open datasets that prioritize educational content, reasoning tasks, and mathematical problem-solving. As mentioned earlier, recent studies have shown that the gradual introduction of high-quality data via a curriculum or multi-stage training can enhance model performance, particularly on reasoning and knowledge-intensive tasks. Given that our datasets do not contain a significant amount of such content, we hypothesized that supplementing them with English datasets focused on these areas could provide beneficial cross-lingual transfer effects. 

For this stage of training, we curate a high-quality dataset by applying quality-based filtering criteria. For Hindi, we retain only documents with an educational score of 3 or higher, ensuring that the selected data contains substantial informational content. For English data, we incorporate text from several high-quality sources. These include FineWeb-Edu ($\text{edu\_score} \geq 3$) \citep{penedo2025fineweb2}, which consists of content extracted from educational web pages, and Cosmopedia \citep{benallal2024smollmcorpus}, a synthetic dataset comprising textbooks, blog posts, stories, WikiHow articles, and other educational-style content. We further include OpenScience \citep{nvidia_openscience} and Big-Reasoning-Traces \citep{allenai_bigreasoningtraces}, which provide data designed to improve reasoning abilities across science, law, economics, and the humanities. Additionally, we incorporate FineMath \citep{allal2025smollm2} and the AllenAI math-meta-reasoning \citep{allenai_math_meta} datasets, which contain high-quality mathematical and educational content.
Table \ref{tab:multi-stage-dataset} summarizes the datasets used in this recipe.

\begin{table}[h]
\centering
\caption{Summary of Datasets Used in Multi-Stage Recipe}
\label{tab:multi-stage-dataset}
\begin{tabular}{@{}l l r@{}}
\toprule
\textbf{Dataset Name} & \textbf{Subset} & \textbf{Size (Tokens)} \\
\midrule

\textbf{\href{https://huggingface.co/datasets/Polygl0t/gigalekh-v1}{\gigalekh}} &  & 50B (Total) \\
 & Edu Score of 3 & 34.95B \\
 & Edu Score of 4 & 14.30B \\
 & Edu Score of 5 & 163.96M \\

\midrule
\href{https://huggingface.co/datasets/HuggingFaceFW/fineweb-edu}{HuggingFaceFW/fineweb-edu} &  & 49.29B (Total) \\
 & Edu Score of 3 & 35.00B \\
 & Edu Score of 4 & 14.22B \\
 & Edu Score of 5 & 69.61M \\

\midrule
\href{https://huggingface.co/datasets/HuggingFaceTB/finemath}{HuggingFaceTB/finemath} &  & 9.66B (Total) \\
 & Edu Score of 4 & 8.59B \\
 & Edu Score of 5 & 1.08B \\

\midrule
\href{https://huggingface.co/datasets/HuggingFaceTB/smollm-corpus}{HuggingFaceTB/smollm-corpus} & All & 25.0B \\
\href{https://huggingface.co/datasets/allenai/big-reasoning-traces}{allenai/big-reasoning-traces} & All & 2.44B \\
allenai/math-meta-reasoning-filtered & All & 1.24B \\
\href{https://huggingface.co/datasets/nvidia/OpenScience}{nvidia/OpenScience} & All & 9.87B \\

\bottomrule
\end{tabular}
\end{table}

With these datasets, we designed a multi-stage training curriculum consisting of three distinct phases: (1) Warmup+Stable, (2) Stable, and (3) Stable+Decay. Each stage focuses on a different data mixture, in which we progressively up-sample higher-quality subsets. Throughout these stages, we keep the language mix approximately balanced(\(\approx 50{:}50\) Hindi: English) across all three stages. Table~\ref{tab:language-proportion-hi} summarizes the language proportions of each stage.

\begin{table}[h]
\centering
\caption{Language Proportions Across Stages in Multi-Stage Recipe}
\label{tab:language-proportion-hi}
\begin{tabular}{@{}l r r@{}}
\toprule
\textbf{Stage} & \textbf{Hindi\% (tokens)} & \textbf{English\% (tokens)} \\
\midrule
Warmup+Stable & 48.5 ($\sim$70B) & 51.5 ($\sim$74B) \\
Stable & 52 ($\sim$81B) & 48 ($\sim$75B) \\
Stable+SqrtDecay & 53 ($\sim$44B) & 47 ($\sim$39B) \\
\bottomrule
\end{tabular}
\end{table}

How many times each portion of our data mixture was repeated during training was designed to yield a total training volume of approximately 356 billion tokens, without excessively over-sampling any particular subset. To achieve this, we tuned the repetition factor $R_D$, defined as $ \operatorname{max}\left( \frac{D}{U} - 1, 0 \right)$
where $D$ is the total number of training tokens and $U$ is the number of unique tokens, keeping it below the regime where diminishing returns from repeated data become significant \citep{muennighoff2023scaling}. 

Concretely, for the multi-stage recipe, we followed a selective token-use strategy. We used only about half of the Hindi tokens from the first recipe, specifically the high-quality subset with \texttt{edu\_score} $\geq 3$, and paired them with an approximately equal number of English tokens for training. This setup enabled us to assess whether prioritizing higher-quality Hindi data yields measurable performance gains. A detailed breakdown of the data mixtures for each stage is provided in Appendix~\ref{appendix:data-mixture}.

Each stage employs a specific moment of a "trapezoidal" learning rate \citep{hagele2024scaling}, also referred to as Warmup-Stable-Decay (WSD) \citep{hu2024minicpm}. Studies such as those by \citep{allal2025smollm2} and \citep{bakouch2025smollm3} have shown that this style of learning rate schedule can promote stable convergence and improved generalization, especially in multi-stage training setups without predetermined epoch boundaries.

We use a linear warmup of 2,000 (1,000 fewer than v0.1) optimization steps to reach the peak learning rate of $7e-4$, followed by a 147,000-step stable phase and a 21,000-step cooldown that follows a negative-square-root shape rather than a linear decay. We chose the negative square-root cooldown because recent works \citep{dremov2025training,hagele2024scaling,apertus2025apertus} show that this shape provides a better balance between continued loss-landscape exploration and conservative annealing during the final phase, often yielding more reliable final convergence than a linear cooldown. In terms of total compute, the multi-stage recipe requires  $\sim1.285 \times 10^21$ FLOPs for the entire training.

\newpage
\section{Pretraining Results}
\label{sec:training-results}

\subsection{Learning Curves and Gradient Statistics}

\begin{figure}[h]
\centering
\includegraphics[width=0.95\linewidth]{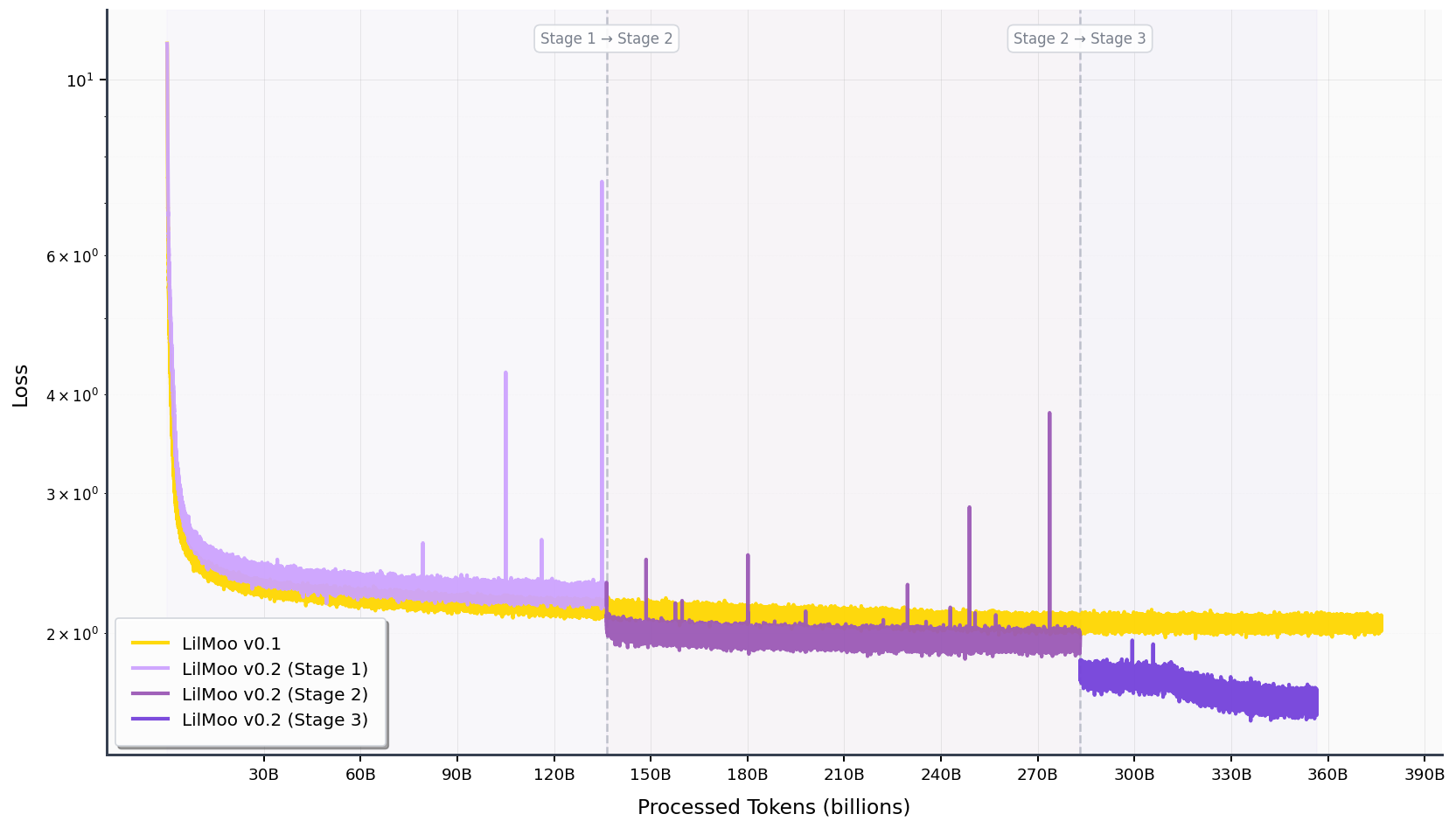}
\caption{Pretraining loss curve for the LilMoo pair.}
\label{fig:pretraining-loss-curve-hi}
\end{figure}

Figure~\ref{fig:pretraining-loss-curve-hi} displays the training loss curves of \lilmoohi (179,590 steps, $\sim$376B tokens)) and \lilmoohien (170,000 steps, $\sim$356B tokens) corresponding to the single-stage (v0.1) and multi-stage (v0.2) training recipes. For both recipes, pretraining was stable, with no major loss spikes or rollbacks. While the simple recipe shows steady improvement over time, the multi-stage recipe exhibits discontinuous loss jumps throughout training, indicative of our changes in data mixtures, which lead to differences in average cross-entropy. However, these discontinuous junctures are to be expected in multi-stage training setups \citep{zhang2024tinyllama,apertus2025apertus}.

\begin{figure}[!htbp]
\centering
\begin{tabular}{cc}
\includegraphics[width=0.48\linewidth]{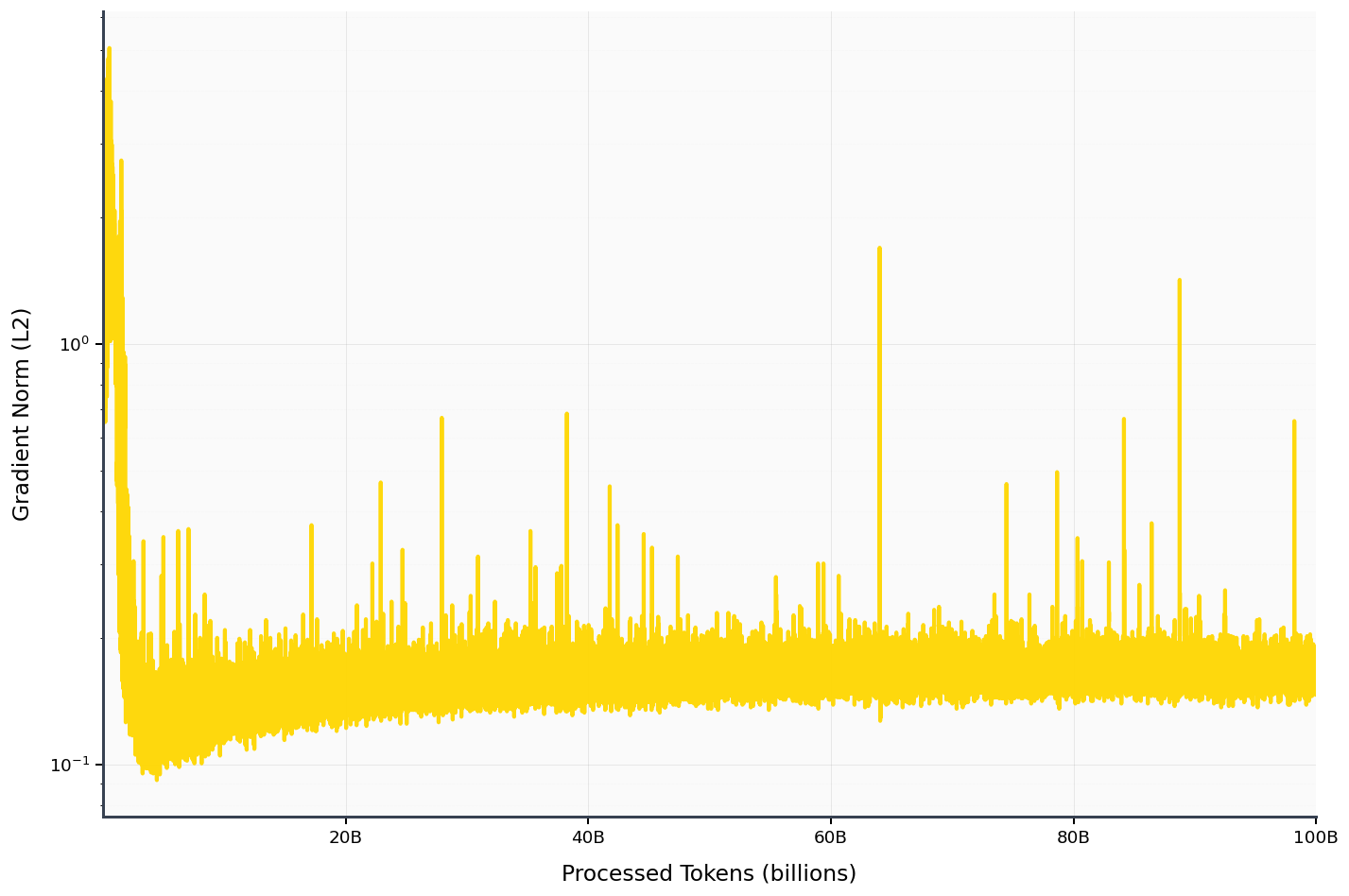} &
\includegraphics[width=0.48\linewidth]{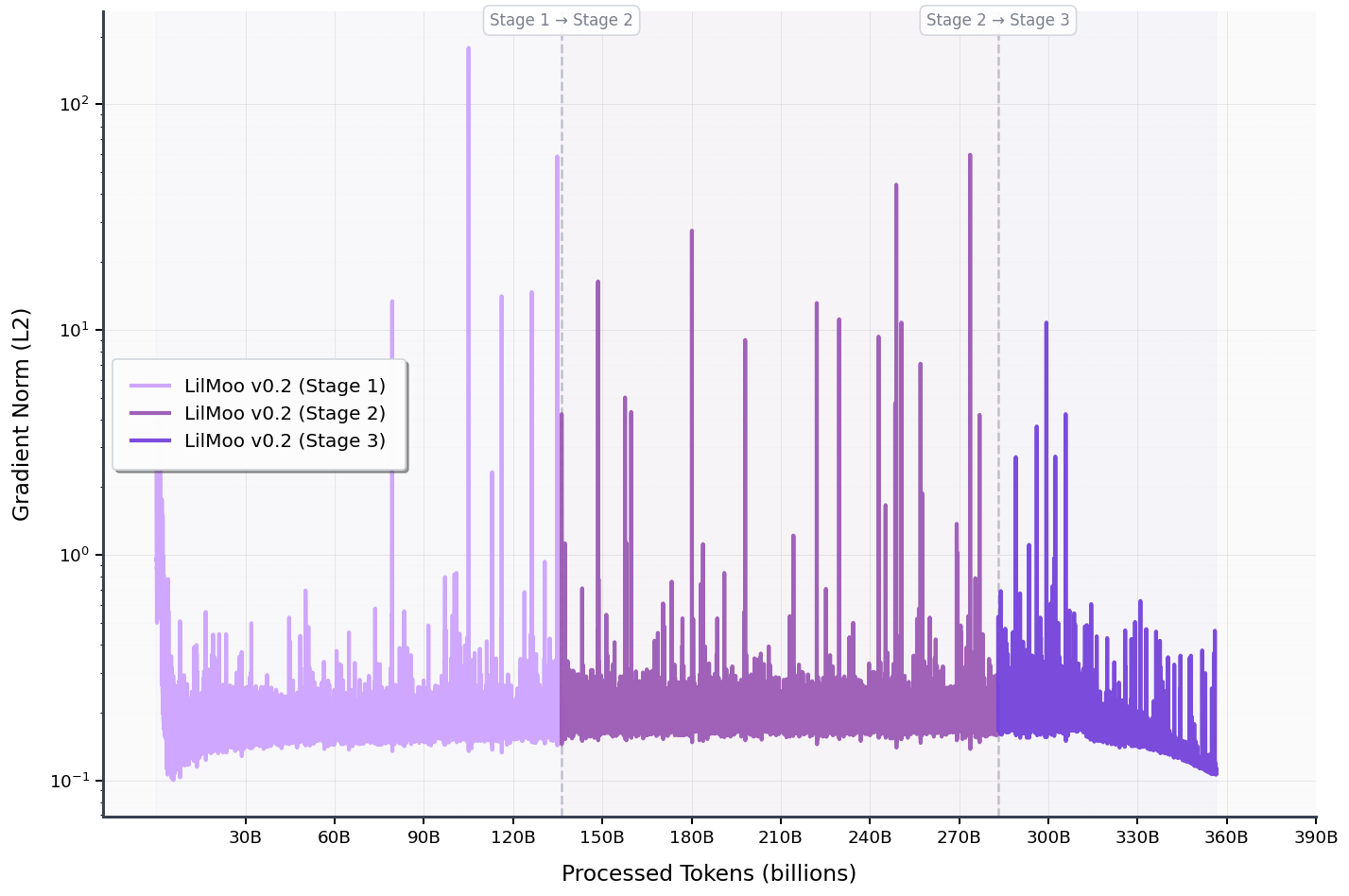} \\
(a) Gradient Statistics (\lilmoohi) & (b) Gradient Statistics (\lilmoohien) \\
\end{tabular}
\caption{Gradient Statistics for the LilMoo pair.}
\label{fig:gradient-statistics-hi}
\end{figure}

The gradient statistics in Figure~\ref{fig:gradient-statistics-hi} further display differences in the training stability of the two recipes. These logs are available in each model's respective repository.

\subsection{Per-benchmark Performance}

Before presenting results on the selected evaluation benchmarks, we conducted an initial analysis to identify which benchmarks provide meaningful signals of model progress, particularly for smaller models like \lilmoo that are trained for shorter durations. This analysis focused on the first version of our monolingual Hindi model (\lilmoohi) and evaluated each Hindi-specific benchmark individually. Benchmarks that demonstrated early sensitivity to improvements in model performance were prioritized for the main evaluations. This approach ensures that the selected benchmarks are both informative and representative for comparing subsequent model versions (\lilmoohi vs. \lilmoohien). Detailed results of this analysis are provided in Appendix~\ref{appendix:evaluation-benchmarks}.

As a baseline for comparison, we selected models from the Qwen series \citep{yang2025qwen3}, due to (1) their strong performance on multilingual benchmarks, and (2) the fact that they provide models with a comparable parameter count to our own models. According to their documentation, both models support Hindi text input and generation. Given that this work focuses solely on pretraining, we did not use the instruction-tuned variants of these models in our evaluations/comparisons. The metric used to report performance on these benchmarks is normalized accuracy.

\begin{figure}[h]
\centering

\begin{subfigure}[b]{0.48\textwidth}
    \centering
    \includegraphics[width=\textwidth]{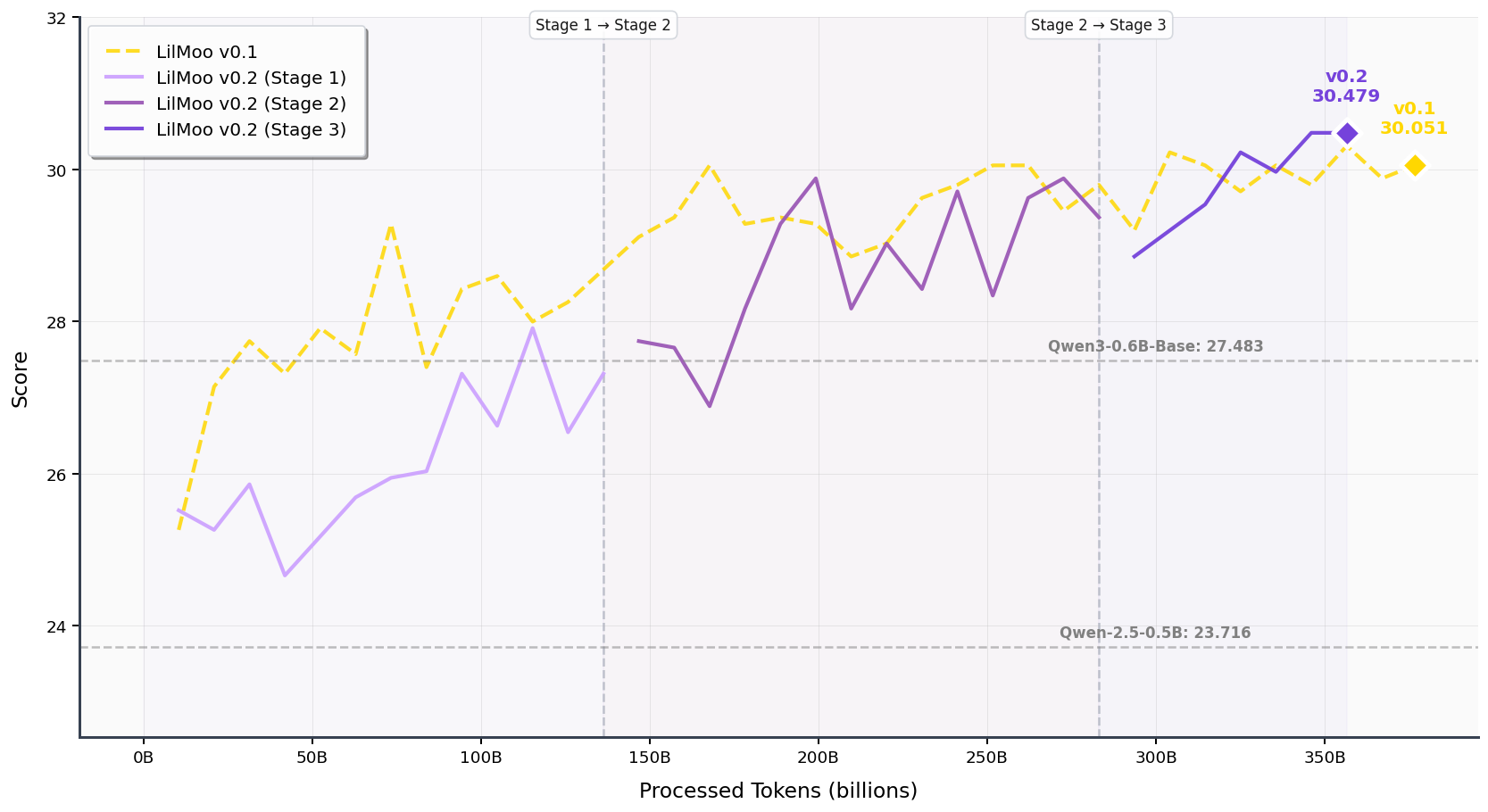}
    \caption{ARC}
    \label{fig:result-arc}
\end{subfigure}
\hfill
\begin{subfigure}[b]{0.48\textwidth}
    \centering
    \includegraphics[width=\textwidth]{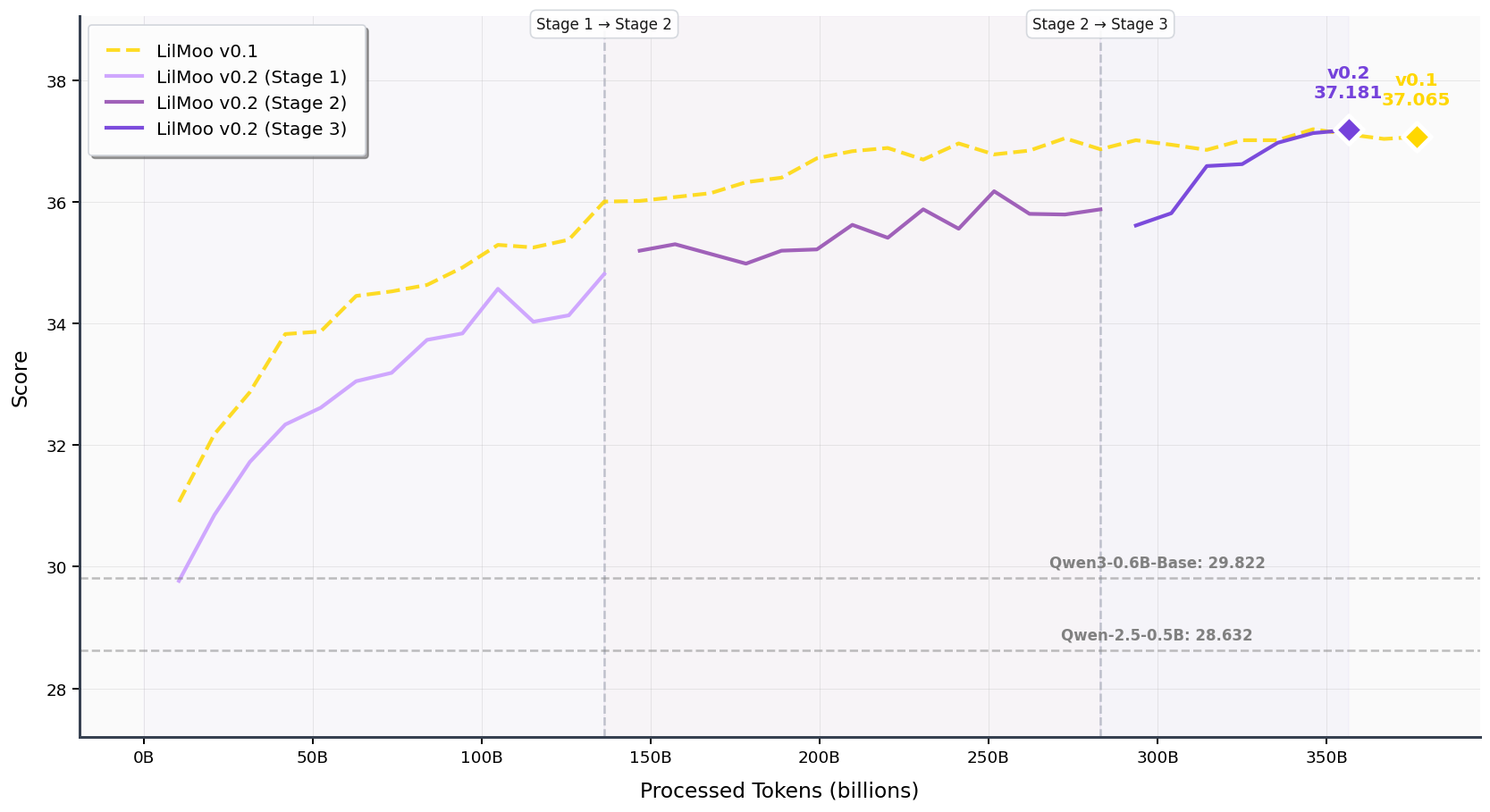}
    \caption{Hellaswag}
    \label{fig:result-hellaswag}
\end{subfigure}

\vspace{0.5cm}

\begin{subfigure}[b]{0.48\textwidth}
    \centering
    \includegraphics[width=\textwidth]{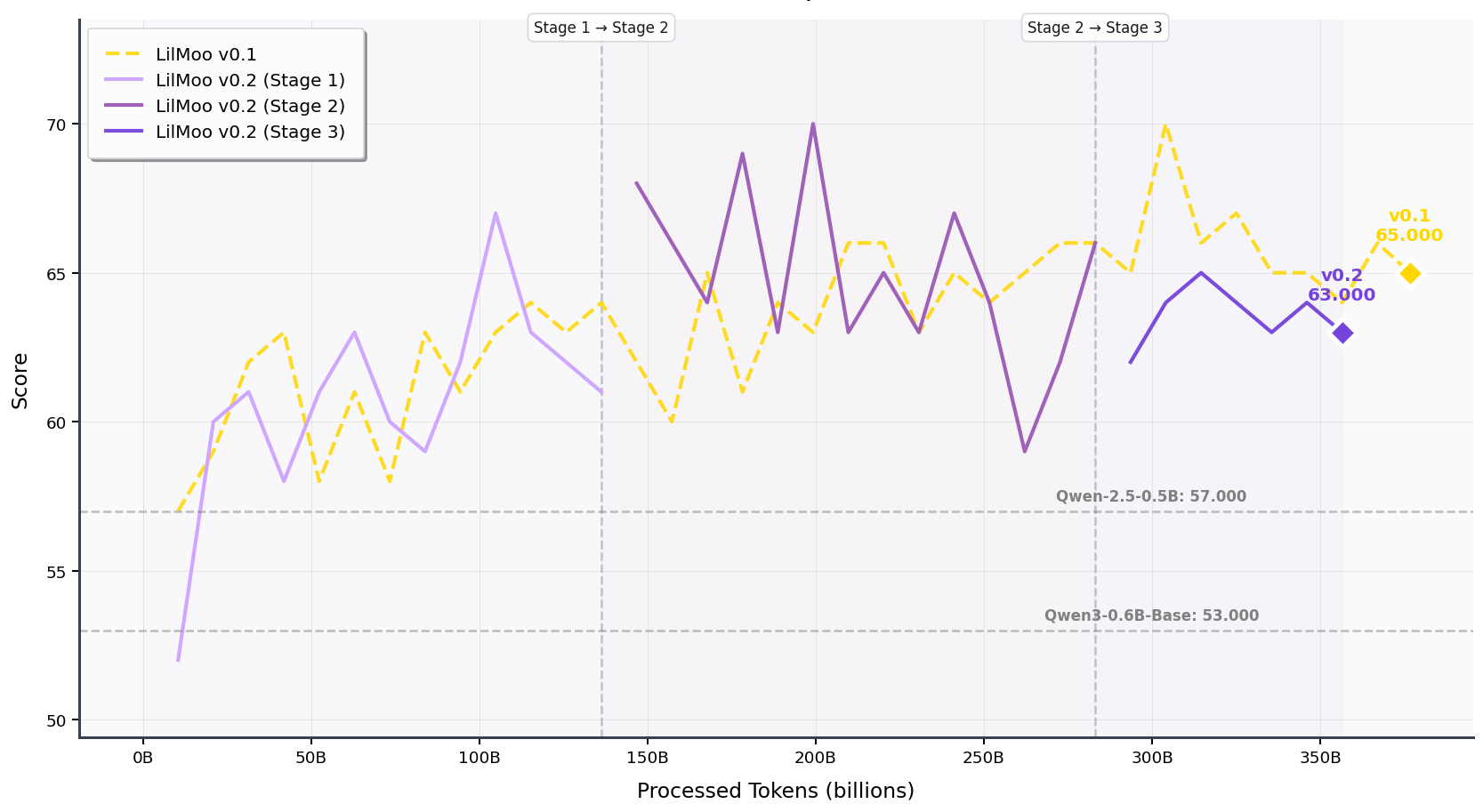}
    \caption{Global PIQA}
    \label{fig:results-piqa}
\end{subfigure}
\hfill
\begin{subfigure}[b]{0.48\textwidth}
    \centering
    \includegraphics[width=\textwidth]{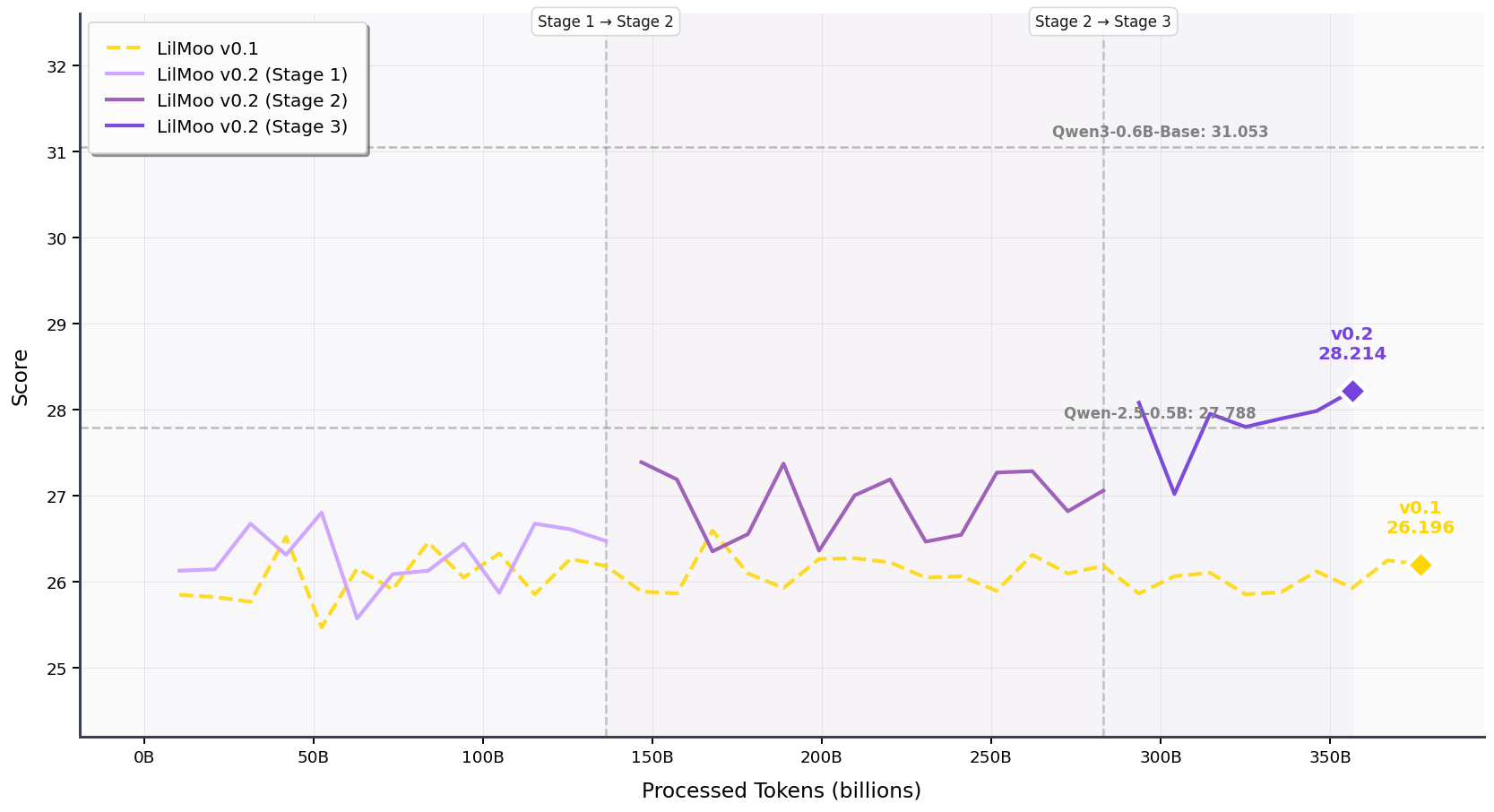}
    \caption{MMLU}
    \label{fig:result-mmlu}
\end{subfigure}

\vspace{0.5cm}

\begin{subfigure}[b]{0.48\textwidth}
    \centering
    \includegraphics[width=\textwidth]{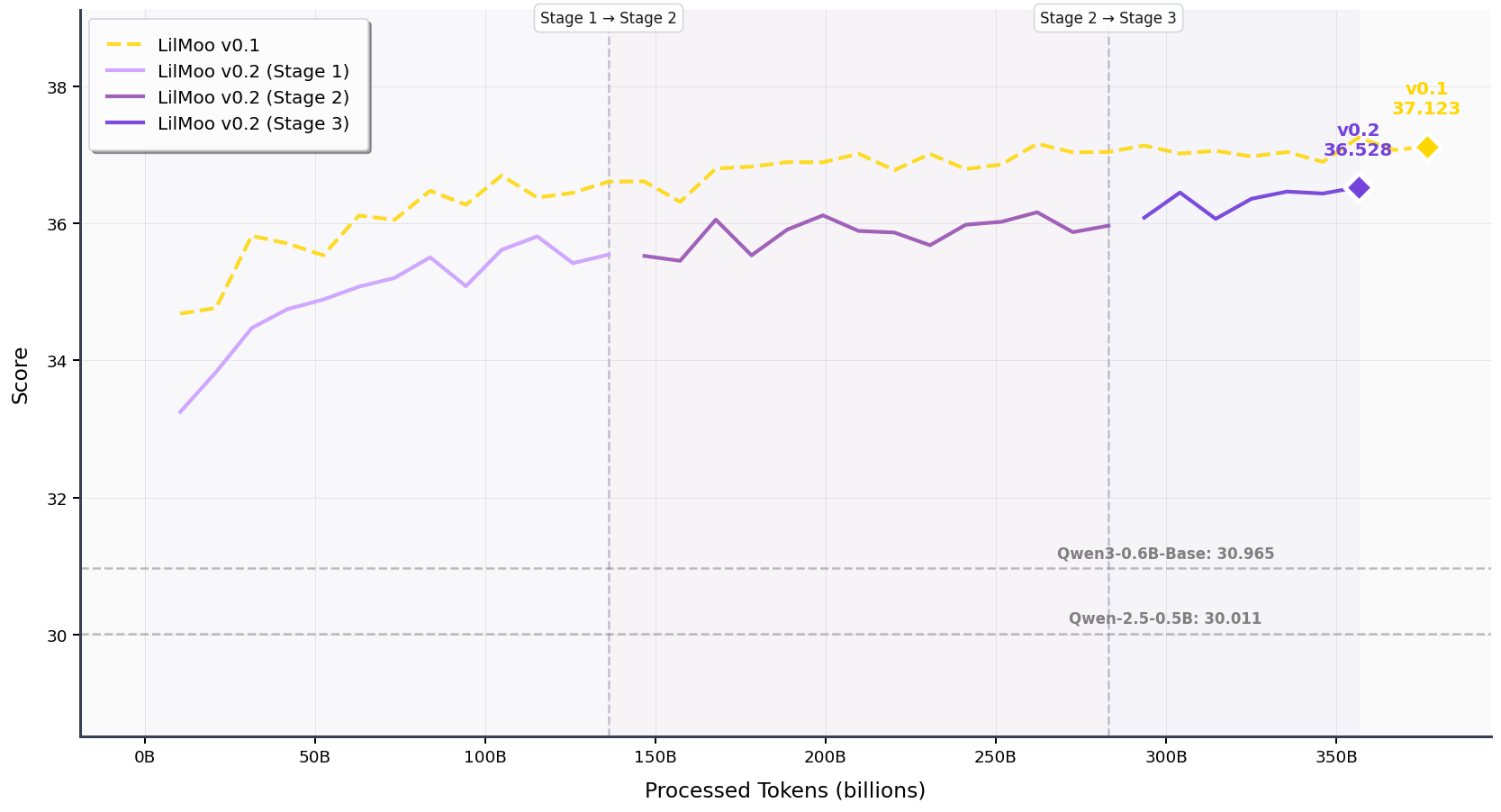}
    \caption{CSQA}
    \label{fig:result-csqa}
\end{subfigure}
\hfill
\begin{subfigure}[b]{0.48\textwidth}
    \centering
    \includegraphics[width=\textwidth]{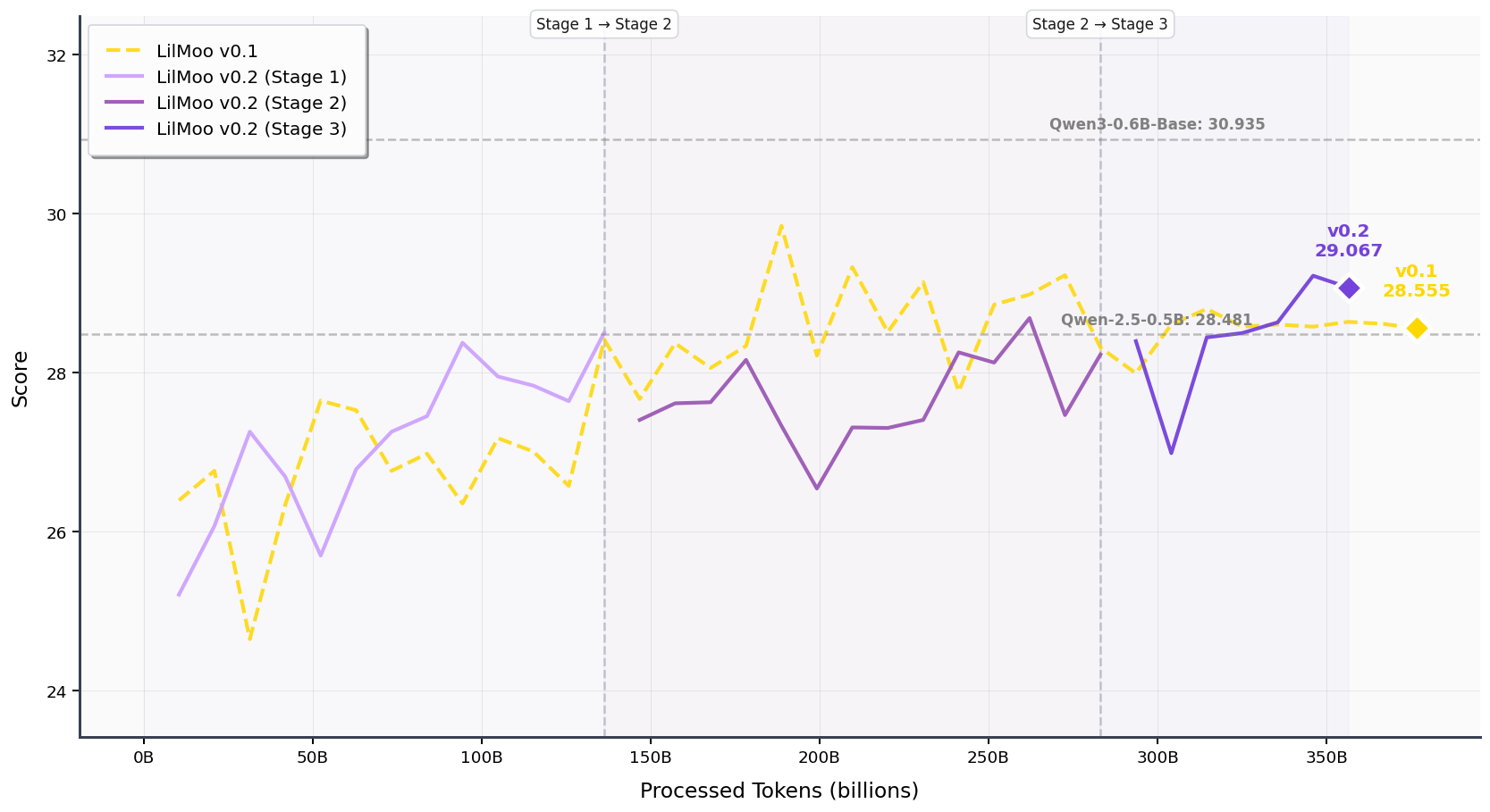}
    \caption{MILU}
    \label{fig:result-milu}
\end{subfigure}

\caption{Per-benchmark performance across our evaluation suite (Qwen2.5/3 serve as Baselines).}
\label{fig:benchmarks-hi}
\end{figure}

Figure \ref{fig:benchmarks-hi} shows how \lilmoo performs across our evaluation suite during pretraining. Results vary considerably by benchmark. ARC, HellaSwag, and CSQA improve steadily as training progresses, whereas Global PIQA and MILU exhibit greater variability but still trend upward overall. MMLU exhibits the greatest instability, with large fluctuations particularly for \lilmoohi. Overall, \lilmoohien generally demonstrates more consistent progress than \lilmoohien.

Across all tasks, \lilmoohien outperforms \lilmoohi on most benchmarks. Both variants substantially outperform baselines (Qwen2.5/3) on ARC, HellaSwag, and CSQA, though the performance gap between them is small on some tasks. On MMLU, Qwen3 significantly outperforms all others; however, \lilmoohien still achieves the second-best performance, outperforming Qwen2.5 by a moderate margin despite being trained on substantially less data. For Global PIQA, the monolingual Hindi model (\lilmoohi) clearly outperforms the bilingual model (\lilmoohien). Given that Global PIQA is a highly culturally specific task that covers local customs, traditions, and everyday life of the native speakers, this result may indicate that the inclusion of English data may hurt the model’s understanding of culture-specific context, thereby reducing performance. Overall, both \lilmoo models consistently outperform Qwen2.5-0.5B and Qwen-3-0.6B-Base, surpassing them on five out of six benchmarks.

\subsection{Aggregate Performance}

\begin{figure}[hbt!]
    \centering
    \includegraphics[width=\textwidth]{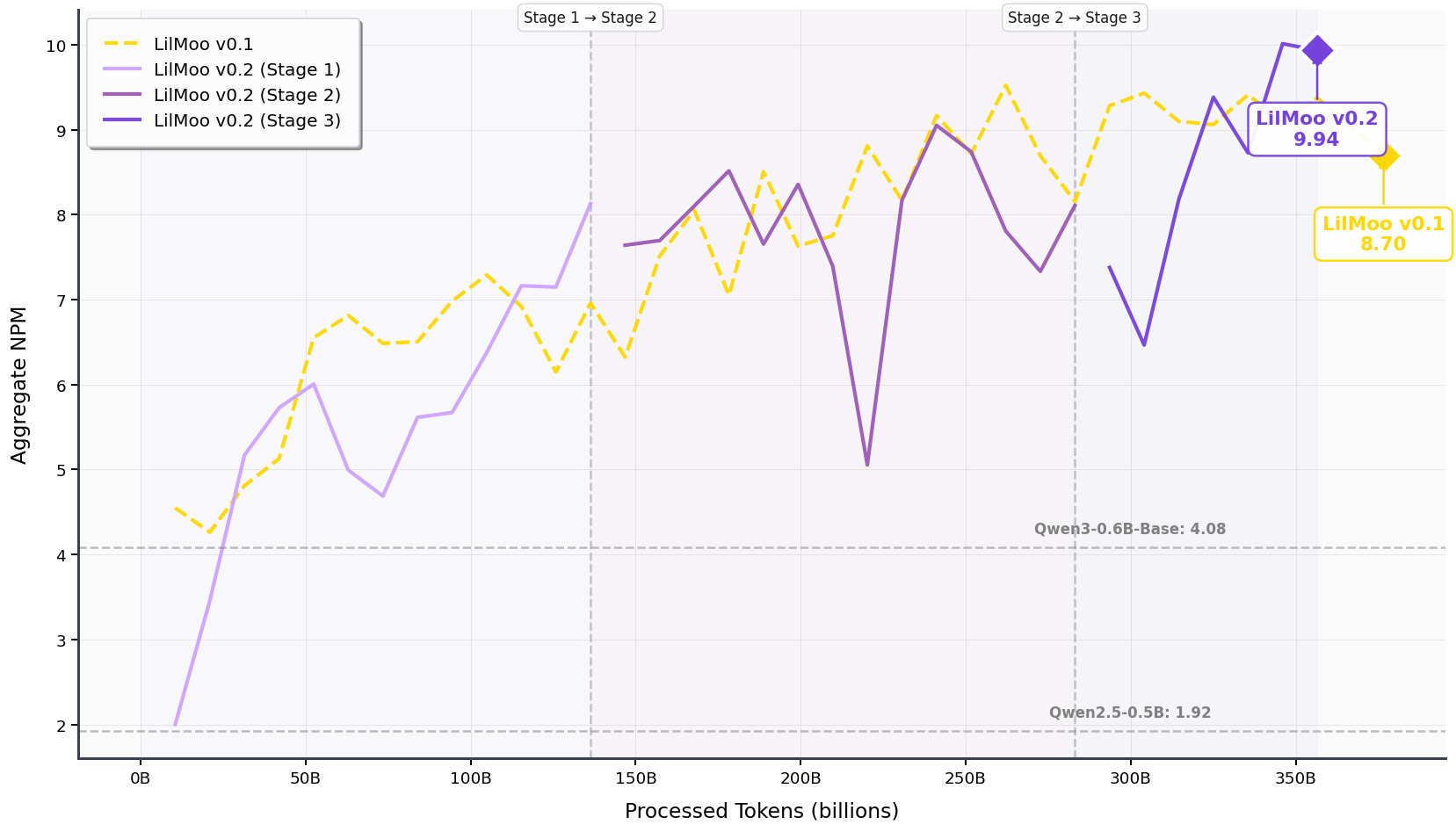}
    \caption{NPM scores for both LilMoo models and both Qwen baselines.}
    \label{fig:npm-final}
\end{figure}

For an overall summary of model performance across all benchmarks, we use the Normalized Preferred Metric (NPM) \citep{pires2023sabia}. NPM provides a single aggregated score by normalizing each benchmark according to its baseline and maximum achievable performance. A full definition of this normalization procedure is provided in Appendix~\ref{appendix:npm}. The NPM score was computed using accuracy.

In Figure \ref{fig:npm-final}, we show our models' NPM (mean) scores against the Qwen baselines. Both the single-stage (\lilmoohi) and multi-stage (\lilmoohien) models are trained with nearly identical compute budgets. Despite this, the multi-stage model still achieves a clear improvement, increasing the aggregate NPM score from $8.70$ (\lilmoohi) to $9.94$ (\lilmoohien). Both \lilmoo variants significantly outperform the Qwen3-0.6B-Base ($4.08$) and Qwen2.5-0.5B ($1.92$) baselines.

\subsection{Performance Against Other Models}
\label{subsec:perf-larger-models}

In the table \ref{tab:larger-model-comparison}, we compare the performance of \lilmoo models against other models on our Hindi evaluation suite. All values presented in the table represent accuracy metrics, both per benchmark and in aggregate (NPM). Despite being significantly smaller, both our models consistently outperform these larger models across several benchmarks.

\begin{table}[h]
\centering
\caption{Performance comparison against other models, beyond the Qwen2.5/3 baselines.}
\label{tab:larger-model-comparison}
\resizebox{\textwidth}{!}{%
\begin{tabular}{l c c c c c c c}
\toprule
\textbf{Model} & \textbf{Aggregate NPM} & \textbf{MILU} & \textbf{CSQA} & \textbf{ARC} & \textbf{MMLU} & \textbf{HellaSwag} & \textbf{PIQA} \\
\midrule
\lilmoohien & \textbf{9.94} & 0.28 & 0.42 & 0.29 & 0.27 & 0.32 & 0.58 \\
\lilmoohi & 8.70 & 0.28 & 0.42 & 0.25 & 0.26 & 0.32 & 0.58 \\
Qwen3-1.7B & 7.80 & 0.37 & 0.34 & 0.26 & 0.36 & 0.28 & 0.50 \\
OpenHathi-7B-Hi-v0.1-Base & 7.64 & 0.30 & 0.36 & 0.23 & 0.27 & 0.31 & 0.58 \\
Airavata & 6.63 & 0.28 & 0.36 & 0.22 & 0.27 & 0.29 & 0.58 \\
gemma-2-2b-it & 6.53 & 0.36 & 0.36 & 0.22 & 0.31 & 0.29 & 0.50 \\
gemma-3-1b-it & 4.80 & 0.29 & 0.36 & 0.21 & 0.28 & 0.29 & 0.52 \\
gemma-3-1b-pt & 4.28 & 0.25 & 0.39 & 0.22 & 0.25 & 0.30 & 0.52 \\
Qwen3-0.6B-Base & 4.08 & 0.30 & 0.33 & 0.23 & 0.31 & 0.28 & 0.49 \\
Llama-3.2-1B-Instruct & 4.07 & 0.28 & 0.36 & 0.23 & 0.28 & 0.29 & 0.49 \\
gemma-2-2b & 3.96 & 0.33 & 0.37 & 0.20 & 0.24 & 0.26 & 0.51 \\
Qwen2.5-1.5B & 3.86 & 0.30 & 0.33 & 0.22 & 0.31 & 0.28 & 0.48 \\
Llama-3.2-1B & 3.35 & 0.27 & 0.37 & 0.23 & 0.26 & 0.29 & 0.48 \\
Qwen-2.5-1.5B-Instruct & 3.26 & 0.30 & 0.33 & 0.23 & 0.31 & 0.28 & 0.46 \\
\bottomrule
\end{tabular}%
}
\end{table}

\newpage
\section{Resource Consumption}
\label{subsec:resource}

An interesting aspect to consider is the relationship between the amount of resources required to train a model and its resulting performance. Resource consumption can be measured in various ways, such as energy consumption or carbon emissions. To monitor/estimate these values, we used the CodeCarbon \citep{codecarbon} library during our training runs. Table \ref{tab:npm-final} summarizes the training duration, energy consumed, and CO$_2$eq emissions, among other compute-related statistics, for both versions/recipes we experimented with.

\begin{table}[h]
\centering
\caption{Training Resource Consumption}
\label{tab:npm-final}
\resizebox{\textwidth}{!}{%
\begin{tabular}{l c c c c c}
\toprule
\textbf{Model} & \makecell{\textbf{Total Duration} \\ \textbf{(hours)}} & \textbf{GPU-hours} & \makecell{\textbf{Energy consumed} \\ \textbf{(kWh)}} & \makecell{\textbf{CO$_2$ emitted} \\ \textbf{(kg CO$_2$eq)}} & \makecell{\textbf{Total Compute} \\ \textbf{(FLOPs)}} \\
\midrule
LilMoo-v0.2 & $\sim$327.50 & $\sim$2648.7 & 1346.76 & 513.05 & $\sim$1.285e21 \\
LilMoo-v0.1 & $\sim$346.97 & $\sim$2797.1 & 1414.21 & 538.74 & $\sim$1.357e21 \\
\bottomrule
\end{tabular}%
}
\end{table}

\begin{figure}[hbt!]
    \centering
    \includegraphics[width=\textwidth]{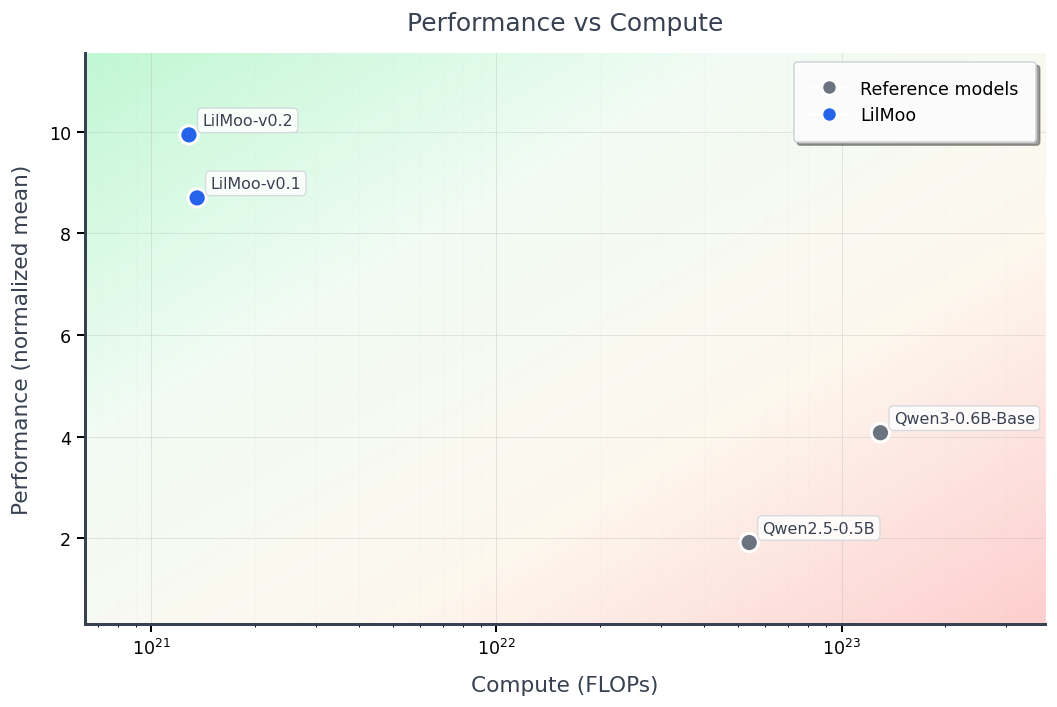}
    \caption{Performance vs Compute for LilMoo Models vs Qwen Baselines}
    \label{fig:perf-vs-comp}
\end{figure}

Figure \ref{fig:perf-vs-comp} reveals that both \lilmoohi and \lilmoohien achieve strong NPM (mean) scores of $8.70$ and $9.94$, respectively, while using approximately $1.357e21$ and $1.285e21$ FLOPs. In contrast, Qwen2.5-0.5B and Qwen3-0.6B-Base require approximately $5.400e22$ and $1.296e23$ FLOPs, respectively, despite achieving lower NPM scores. To put these values into perspective, if we compute the ratio of FLOPs required by Qwen3-0.6B-Base to that of our multi-stage models, we find that Qwen3-0.6B-Base uses $\sim$100 times more compute than \lilmoo models to achieve a lower NPM (mean) score. In other words, if our methods yielded similar results across other low-resource languages, this would imply that the compute budget required to train a single large multilingual model such as Qwen3-0.6B-Base could instead be used to develop $\sim$100 language-specific models\footnote{If data were available.} of similar size that achieve better overall performance.

\newpage
\section{Limitations and Future Work}
\label{sec:limitations-future-work}

During our experiments, we identified several areas for potential improvement and further investigation that we were unable to explore fully within the scope of this work, given that resources for extensive experimentation and ablation studies are non-trivial to obtain. Hence, we list, in this section, interesting areas to explore in future work.

\paragraph{Evaluation of data pipelines.} Future work could compare different data pipelines (e.g., JQL \cite{ali2025judging}) against the FineWeb2 pipeline. Such experiments may reveal which dataset curation strategies are most effective for low-resource languages, such as Hindi.

\paragraph{Scaling model sizes.} We focused on 0.5B–0.6B parameter models to stay within compute budget constraints and enable faster experimentation. Future work should explore larger models (e.g., 1B–3B parameters) trained with similar recipes to determine whether the observed advantage over Qwen2.5/3 baselines persists at larger scales.

\paragraph{Training curriculum ablations.} Our multi-stage training recipe was not fully explored. Systematic ablation studies could identify optimal curriculum designs for low-resource language modeling, considering factors such as language proportions, data mixtures, and compute budgets (e.g., 1e20 FLOPs).

\paragraph{Cross-lingual transfer analysis.} While our multi-stage recipe shows promising language transfer effects, it remains to be tested extensively across other low-resource languages, including additional Indic languages (e.g., Marathi, Tamil, Telugu) and languages from different families.

Despite these unexplored directions, our work establishes a strong foundation for future research in Hindi. For instance, while initiatives such as the Goldfish models \citep{chang2024goldfish} have made meaningful progress by developing monolingual models for low-resource languages, their systems remain relatively small (125M parameters), undertrained (only 9e17 FLOPs), and consequently limited in capability. As a result, they fall short of the performance levels necessary to ``perform'' on most commonly used evaluations. In contrast, our preliminary results demonstrate substantially higher performance. If these recipes yield comparable results across other low-resource languages (as stated in Section~\ref{subsec:resource}), the compute budget of a single Qwen3-0.6B-Base model could instead train nearly 100 specialized, language-specific models---potentially accelerating progress in low-resource language modeling across a wide spectrum of languages.

\newpage
\section{Conclusion}
\label{sec:conclusion}

In this work, we present \lilmoo, a pair of compact language models for Hindi, trained using both single-stage and multi-stage approaches. Our results show that general language understanding can be achieved for low-resource languages through carefully designed training curricula and targeted exposure to monolingual and bilingual-augmented data. Both versions of our models consistently outperformed Qwen baselines, despite being trained with significantly less compute. We further observe that selectively including high-quality English data can enhance performance: the bilingual Hindi–English model outperformed the Hindi-only model on most benchmarks. At the same time, incorporating additional languages can negatively affect culturally grounded tasks, such as Global PIQA, underscoring the importance of balancing and curating bilingual training. Adjusting language ratios or exploring alternative curriculum strategies may yield further improvements. While our experiments provide evidence that compact, targeted models can achieve strong performance on low-resource languages, the generalization of these findings to other languages (and scales) remains uncertain. Conducting further controlled studies across model sizes, curriculum strategies, and dataset scales would be valuable for fully validating our observations. Overall, this work provides evidence that effective low-resource language modeling is achievable with limited compute and careful data curation at small scales, providing a practical foundation for future research on compact, performant models for low-resource languages.

\newpage
\section*{Acknowledgements}

Polyglot is a project funded by the Federal Ministry of Education and Research (BMBF) and the Ministry of Culture and Science of the State of North Rhine-Westphalia (MWK) as part of TRA Sustainable Futures (University of Bonn) and the Excellence Strategy of the federal and state governments.

A.S. acknowledges funding by the Deutsche Forschungsgemeinschaft (DFG, German Research Foundation) as part of the CRC 1639 NuMeriQS – project No. 511713970.

We also gratefully acknowledge the granted access to the \href{https://www.hpc.uni-bonn.de/en/systems/marvin}{Marvin cluster} hosted by \href{https://www.uni-bonn.de/en}{University of Bonn} along with the support provided by its High Performance Computing \& Analytics Lab.

This work was completed as part of the Master’s thesis of Shiza Fatimah at the University of Bonn, under the supervision and guidance of Lucie Flek, Nicholas Kluge Corrêa, and Florian Mai

\newpage
\section*{Authors Contribution}

The corresponding author is \textbf{Shiza Fatimah}. She is a master's student working at the Bonn-Aachen International Center for Information Technology (b-it) / CAISA Lab, as part of the Lamarr Institute for Machine Learning and Artificial Intelligence, University of Bonn (Bonn, NRW, Germany). Her contact email is \href{mailto:s39sfati@uni-bonn.de}{s39sfati@uni-bonn.de}. \textbf{Sh.F.} contributed to the project's idealization, development of the software stack, dataset curation, training, and evaluation of the models, as well as writing the article and documenting the repositories.

\textbf{Nicholas Kluge Corrêa} is a postdoc researcher at the Bonn-Aachen International Center for Information Technology (b-it) / CAISA Lab, as part of the Lamarr Institute for Machine Learning and Artificial Intelligence, University of Bonn (Bonn, NRW, Germany). His contact email is \href{mailto:kluge@uni-bonn.de}{kluge@uni-bonn.de}. \textbf{N.K.C.} contributed to the development of the software stack, as well as writing the article and documenting the repositories. \textbf{N.K.C.} is one of the Principal Investigators of the Polyglot Project.

\textbf{Aniket Sen} is a postdoc researcher at the Helmholtz-Institut für Strahlen und Kernphysik, University of Bonn, and the Bethe Center for Theoretical Physics, University of Bonn (Bonn, NRW, Germany). His contact email is \href{mailto:sen@hiskp.uni-bonn.de}{sen@hiskp.uni-bonn.de}. \textbf{A.S.} contributed to the optimization of the software stack, as well as the article's writing. \textbf{A.S.} is one of the Principal Investigators of the Polyglot Project.

\textbf{Sophia Falk} is a PhD researcher at the Bonn Sustainable AI Lab, Institute for Science and Ethics, University of Bonn (Bonn, NRW, Germany). Her contact email is \href{mailto:falk@iwe.uni-bonn.de}{falk@iwe.uni-bonn.de}. \textbf{So.F.} contributed to implementing the carbon tracking methodology, monitoring training runs, and writing the article.

\textbf{Lucie Flek} is a full professor at the University of Bonn, leading the Data Science and Language Technologies group. Her contact email is \href{mailto:flek@bit.uni-bonn.de}{flek@bit.uni-bonn.de}. \textbf{L.F.} contributed to the project's idealization, as well as writing and reviewing the article.

\textbf{Florian Mai} is a Junior Research Group Leader of the mAI-alignment group at the Bonn-Aachen International Center for Information Technology (b-it) / CAISA Lab, as part of the Lamarr Institute for Machine Learning and Artificial Intelligence, University of Bonn (Bonn, NRW, Germany). His contact email is \href{mailto:fmai@bit.uni-bonn.de}{fmai@bit.uni-bonn.de}. \textbf{F.M.} contributed to the project's idealization, as well as writing and reviewing the article.

\newpage
\begin{small}
\bibliography{biblio}
\bibliographystyle{iclr2025_conference}
\end{small}

\newpage
\appendix

\setcounter{table}{0}
\setcounter{figure}{0}

\renewcommand\thefigure{\thesection.\arabic{figure}}
\renewcommand\thetable{\thesection.\arabic{table}}

\section*{Appendix}
\section{LLM Judge Prompts}
\label{appendix:llm-judge-prompts}

\subsection{Educational Quality Evaluation Prompt}

The following prompt was used with Qwen/Qwen2.5-32B-Instruct to evaluate the educational quality of documents. The prompt is adapted from the FineWeb-Edu dataset \citep{penedo2024fineweb}:

\begin{tcolorbox}[colback=gray!5!white, colframe=gray!75!black, title=Educational Quality Prompt]
\small
You are a text quality evaluator for Hindi. Your task is to assess, on a scale of 1 to 5, how informative and educational a given text sample is. You MUST output your score in a JSON format.

\textbf{Evaluation Criteria:}

\begin{itemize}
    \item Give 1 if the text is not informative or educational. Also, give 1 if the text is too simple, too short, ill-formatted, non-sensical, or contains NSFW content.
    \item Give 2 if the text is somewhat informative but lacks educational value. For example, it might mix educational content with non-educational material, offering a superficial overview of potentially useful topics.
    \item Give 3 if the text is informative and appropriate for educational use and introduces key concepts relevant to school curricula. For example, if the text is a well-written article on a scientific topic, but it may not be comprehensive or could include some extraneous information, being overly complex or too simple.
    \item Give 4 if the text is educational and informative, providing a highly relevant and beneficial educational purpose for a level not higher than grade school, exhibiting a clear and consistent writing style. For example, it could be similar to a chapter from a textbook or a tutorial, offering substantial educational content, including exercises and solutions, with minimal irrelevant information.
    \item Give 5 if the text is highly educational and informative. For a 5 score, the text must be outstanding in its educational value, perfectly suited for teaching either at primary school or grade school. It follows detailed reasoning, the writing style is easy to follow and offers profound and thorough insights into the subject matter.
\end{itemize}

\textbf{Output Requirements:}

\begin{itemize}
    \item Your output must be a valid JSON object.
    \item The JSON must contain only one key-value pair.
    \item The key must be 'score'.
    \item The value must be an integer between 1 and 5.
    \item The value must be a faithful representation of the criteria above.
\end{itemize}

\textbf{Example Output:}

\texttt{\{'score': 2\}}

Ensure that your responses strictly adhere to this format and evaluate the text accurately based on the given criteria.

Below is an excerpt from a webpage. Rate whether the page has high educational value and can be useful in an educational setting for elementary to high school teaching. Assign a score between 1 and 5. Ensure your score reflects how informative and educational the text is. The text will be provided in Hindi.

---

Texto: [THE TEXT TO BE EVALUATED]

---

Provide your score as an integer between 1 and 5. Answer only in JSON format.
\end{tcolorbox}

\subsection{Toxicity Evaluation Prompt}

The following prompt was used with Qwen/Qwen2.5-32B-Instruct to evaluate the toxicity level of documents:

\begin{tcolorbox}[colback=gray!5!white, colframe=gray!75!black, title=Toxicity Evaluation Prompt]
\small
You are a text quality evaluator for the Hindi language. Your task is to assess, from a scale of 1 to 5, how toxic a given text sample is. You MUST output your score in a JSON format.

\textbf{Evaluation Criteria:}

\begin{itemize}
    \item Give 1 if the text is non-toxic, neutral, and appropriate for all audiences. The text contains no offensive, harmful, or inflammatory content.
    \item Give 2 if the text has minor toxicity, such as mildly inappropriate language, slight rudeness, or subtle negative sentiment, but it is not explicitly harmful or offensive.
    \item  Give 3 if the text contains moderate toxicity, including some insults, offensive language, or inappropriate phrasing, but it is not highly aggressive or severely harmful.
    \item Give 4 if the text is highly toxic, containing strong hate speech, explicit threats, severe insults, or harmful stereotypes. It may also include demeaning or aggressive language targeting individuals or groups.
    \item Give 5 if the text is extremely toxic, exhibiting the highest level of harmful content, such as explicit calls for violence, severe discrimination, threats, or extremely offensive language. This score is reserved for the most dangerous and abusive texts.
\end{itemize}

\textbf{Output Requirements:}

\begin{itemize}
    \item Your output must be a valid JSON object.
    \item The JSON must contain only one key-value pair
    \item The key must be 'score'.
    \item The value must be an integer between 1 and 5.
    \item The value must be a faithful representation of the criteria above.
\end{itemize}

\textbf{Example Output:}

\texttt{\{'score': 2\}}

Ensure that your responses strictly adhere to this format and evaluate the text accurately based on the given criteria.

Below is an excerpt from a webpage. Rate whether the excerpt contains harmful, offensive, or inappropriate language. Assign a score between 1 and 5. Ensure your score reflects the level of toxicity present in the text. The text will be provided in Hindi.

---

Texto: [THE TEXT TO BE EVALUATED]

---

Provide your score as an integer between 1 and 5. Respond only in JSON format.
\end{tcolorbox}

\newpage

\section{Data Sources and Statistics}
\label{appendix:base-data-sources-statistics}

This appendix provides detailed information about the data sources, composition, and statistics of \gigalekh.

\subsection{Data Sources and Licenses}

Table~\ref{tab:gigalekh-sources} lists all data sources used to construct \gigalekh, organized by source type. The corpus aggregates data from Common Crawl snapshots, Hugging Face datasets, and other openly available resources. The cutoff date for included data is December 2025.

\begin{table}[h]
\centering
\caption{Data sources organized by source type.}
\label{tab:gigalekh-sources}
\small
\begin{tabular}{@{}llp{5.5cm}@{}}
\toprule
\textbf{Source Type} & \textbf{Dataset / Crawl} & \textbf{License(s)} \\
\midrule
\multirow{10}{*}{Common Crawl} 
& \href{https://data.commoncrawl.org/crawl-data/CC-MAIN-2025-30/index.html}{CC-MAIN-2025-30} & ODC-By v1.0, CommonCrawl ToU \\
& \href{https://data.commoncrawl.org/crawl-data/CC-MAIN-2025-26/index.html}{CC-MAIN-2025-26} & ODC-By v1.0, CommonCrawl ToU \\
& \href{https://data.commoncrawl.org/crawl-data/CC-MAIN-2025-05/index.html}{CC-MAIN-2025-05} & ODC-By v1.0, CommonCrawl ToU \\
& \href{https://data.commoncrawl.org/crawl-data/CC-MAIN-2024-51/index.html}{CC-MAIN-2024-51} & ODC-By v1.0, CommonCrawl ToU \\
& \href{https://data.commoncrawl.org/crawl-data/CC-MAIN-2023-50/index.html}{CC-MAIN-2023-50} & ODC-By v1.0, CommonCrawl ToU \\
& \href{https://data.commoncrawl.org/crawl-data/CC-MAIN-2022-49/index.html}{CC-MAIN-2022-49} & ODC-By v1.0, CommonCrawl ToU \\
& \href{https://data.commoncrawl.org/crawl-data/CC-MAIN-2021-49/index.html}{CC-MAIN-2021-49} & ODC-By v1.0, CommonCrawl ToU \\
& \href{https://data.commoncrawl.org/crawl-data/CC-MAIN-2020-50/index.html}{CC-MAIN-2020-50} & ODC-By v1.0, CommonCrawl ToU \\
\midrule
\multirow{22}{*}{Hugging Face}

& \href{https://huggingface.co/datasets/allenai/c4}{AllenAI-C4} & ODC-By v1.0 \\
& \href{https://huggingface.co/datasets/HuggingFaceTB/finemath}{FineMath (Translated to Hindi)} & ODC-By v1.0 \\
& \href{https://huggingface.co/datasets/HuggingFaceTB/smollm-corpus}{SmolLM-Corpus (Translated to Hindi)} & ODC-By v1.0 \\

& \href{https://huggingface.co/datasets/statmt/cc100}{CC100} & CommonCrawl ToU \\
& \href{https://huggingface.co/datasets/OdiaGenAI/health_hindi_200}{Health-hindi-200} & CommonCrawl ToU \\

& \href{https://huggingface.co/datasets/HuggingFaceFW/fineweb-2}{FineWeb-2} & ODC-By v1.0, CommonCrawl ToU \\

& \href{https://huggingface.co/datasets/HPLT/HPLT2.0_cleaned}{HPLT2.0} & CC0-1.0 \\
& \href{https://huggingface.co/datasets/djstrong/oscar-small}{Oscar-small} & CC0-1.0 \\

& \href{https://huggingface.co/datasets/zicsx/OSCAR-2301-Hindi-Cleaned}{OSCAR-hindi} & Apache 2.0 \\
& \href{https://huggingface.co/datasets/Tensoic/Bhandara}{Bhandara} & Apache 2.0 \\
& \href{https://huggingface.co/datasets/dnyanesh/HindiMathQuest}{HindiMathQuest} & Apache 2.0 \\
& \href{https://huggingface.co/datasets/KathirKs/fineweb-edu-hindi}{Fineweb-edu-hindi} & Apache 2.0 \\

& \href{https://huggingface.co/datasets/wikimedia/wikipedia}{Wikipedia} & CC-BY-SA-3.0 \\
& \href{https://huggingface.co/datasets/bigscience-data/roots_indic-hi_wikipedia}{Roots-indic-hi} & CC-BY-SA-3.0 \\
& \href{https://huggingface.co/datasets/soketlabs/bhasha-wiki}{Bhasha-wiki} & CC-BY-SA-3.0 \\

& \href{https://huggingface.co/datasets/ganeshjcs/hindi-article-summarization}{Hindi-article} & CC-BY-SA-4.0 \\
& \href{https://huggingface.co/datasets/Davlan/sib200}{SIB-200} & CC-BY-SA-4.0 \\

& \href{https://huggingface.co/datasets/MBZUAI/Bactrian-X}{Bactrian-X} & CC-BY-NC-4.0 \\
& \href{https://huggingface.co/datasets/bigscience-data/roots_indic-hi_indic_nlp_corpus}{Roots-indic-nlp} & CC-BY-NC-4.0 \\

& \href{https://huggingface.co/datasets/csebuetnlp/xlsum}{Xlsum} & CC-BY-NC-SA-4.0 \\

\bottomrule
\end{tabular}
\end{table}

\newpage

\section{Classifier Training Details}
\label{appendix:classifier-training}

\paragraph{Training Setup}
All candidate models were fine-tuned for 20 epochs using a batch size of 256 and a maximum sequence length of 512 tokens. Optimization was performed with AdamW \citep{loshchilov2017decoupled} configured with $\beta_1=0.9$, $\beta_2=0.999$, $\epsilon=10^{-8}$, zero weight decay, and a peak learning rate of $3\times10^{-4}$ that decreased linearly to zero over training without a warmup phase. Model states were saved and evaluated every 1000 update steps, with at most five checkpoints retained per run. Final model selection was based on macro F1 measured on a held-out test set of 20{,}000 examples. Aggregate performance across all evaluated models is reported in Table~\ref{tab:evaluation-classifiers}.

\paragraph{Results}
Across both classification tasks, Hindi-RoBERTa achieved slightly stronger scores than IndicBERT and was therefore chosen as the final model used in our filtering pipeline. When the original five-class labels are mapped into binary categories, performance improves further. For the educational classifier, documents with scores $\geq 3$ are treated as high quality and those with scores $<3$ as low quality. For the toxicity classifier, documents with scores $\leq 3$ are considered acceptable, while those with scores $>3$ are treated as toxic. Under this binary formulation, the educational classifier achieves 91.29\% accuracy with a macro F1 of 0.81, and the toxicity classifier achieves 89.57\% accuracy with a macro F1 of 0.74. Both trained models are publicly released to support reproducibility and downstream research.

\begin{table}[h]
\centering
\caption{Evaluation results for educational-quality and toxicity classifiers on the test set using the original 5-class labels.}
\label{tab:evaluation-classifiers}
\begin{tabular}{@{}l l r r r r@{}}
\toprule
\textbf{Model Name} & \textbf{Task} & \textbf{Precision} & \textbf{Recall} & \textbf{F1 Macro} & \textbf{Accuracy} \\
\midrule
\multirow{2}{*}{HindRoBERTa} 
 & Educational & 0.52 & 0.47 & 0.49 & 0.72 \\
 & Toxicity    & 0.59 & 0.45 & 0.47 & 0.70 \\
\midrule
\multirow{2}{*}{IndicBERT} 
 & Educational & 0.51 & 0.47 & 0.48 & 0.71 \\
 & Toxicity    & 0.56 & 0.43 & 0.44 & 0.69 \\
\bottomrule
\end{tabular}
\end{table}   
\newpage

\section{Detailed Data Mixture: Multi-stage Recipe}
\label{appendix:data-mixture}

\subsection{Stage 1 (Warmup+Stable) Data Mixture}

\textbf{Data composition:}
\begin{itemize}
    \item \textbf{Hindi text:}  $\sim$49\% ($\sim$69.8B tokens)
    \item \textbf{Educational English text:}  $\sim$27\% ($\sim$40B tokens) 
    \item \textbf{Reasoning-focused English text:} $\sim$12\% ($\sim$17B tokens) 
    \item \textbf{Math English text:} $\sim$12\% ($\sim$17B tokens)
    \item \textbf{Learning Rate:} linear warmup for the first 2,000 steps, reaching a peak of $7e-4$. It then remains stable at this peak for the next $65,000$ steps before transitioning to the next stage.
\end{itemize}

\begin{table}[h]
\centering
\caption{Data mixture for Stage-1}
\label{tab:stage1-hi}
\begin{tabular}{@{}l l c c@{}}
\toprule
\textbf{Dataset Name} & \textbf{Subset} & \textbf{Size (Tokens)} & \textbf{Repetition Factor} \\
\midrule
\href{https://huggingface.co/datasets/Polygl0t/gigalekh-v1}{\gigalekh} & Edu Score of 3 & 34.9B & 2 \\
\href{https://huggingface.co/datasets/HuggingFaceFW/fineweb-edu}{HuggingFaceFW/fineweb-edu} & Edu Score of 3 & 40.00B & 1 \\
\href{https://huggingface.co/datasets/HuggingFaceTB/finemath}{HuggingFaceTB/finemath} & Edu Score of 4 & 8.59B & 2 \\
\href{https://huggingface.co/datasets/allenai/big-reasoning-traces}{allenai/big-reasoning-traces} & All & 2.44B & 2 \\
allenai/math-meta-reasoning-filtered$^*$ & All & 1.24B & 2 \\
\href{https://huggingface.co/datasets/nvidia/OpenScience}{nvidia/OpenScience} & All & 9.7B & 1 \\
\bottomrule
\end{tabular}
\textit{* At the time of our writing, ``allenai/math-meta-reasoning-filtered'' is no longe available on the Hub.}
\end{table}

\subsection{Stage 2 (Stable) Data Mixture}

\textbf{Data composition:}

\begin{itemize}
    \item \textbf{Hindi text:}  $\sim$52\% ($\sim$81B tokens)
    \item \textbf{Synthetic English text:} $\sim$18\% ($\sim$29B tokens)
    \item \textbf{Educational English text:} $\sim$9\% ($\sim$14B tokens)
    \item \textbf{Reasoning-focused English text:} $\sim$15\% ($\sim$23B tokens)
    \item \textbf{Math English text:} $\sim$6\% ($\sim$9.5B tokens)
    \item \textbf{Learning Rate:} remains stable at $7e-4$ throughout the $ 70,000$-step training.
\end{itemize}

\begin{table}[h]
\centering
\caption{Data mixture for Stage-2}
\label{tab:stage2-hi}
\begin{tabular}{@{}l l c c@{}}
\toprule
\textbf{Dataset Name} & \textbf{Subset} & \textbf{Size (Tokens)} & \textbf{Repetition Factor} \\
\midrule
\href{https://huggingface.co/datasets/Polygl0t/gigalekh-v1}{\gigalekh} & Edu Score of 3 & 34.9B & 1.5 \\
 & Edu Score of 4 & 14.3B & 2 \\
\href{https://huggingface.co/datasets/HuggingFaceFW/fineweb-edu}{HuggingFaceFW/fineweb-edu} & Edu Score of 4 & 14.22B & 1 \\
\href{https://huggingface.co/datasets/HuggingFaceTB/smollm-corpus}{HuggingFaceTB/smollm-corpus} & All & 29.4B & 1 \\
\href{https://huggingface.co/datasets/HuggingFaceTB/finemath}{HuggingFaceTB/finemath} & Edu Score of 4 & 8.59B & 1 \\
 & Edu Score of 5 & 1.08B & 2 \\
\href{https://huggingface.co/datasets/allenai/big-reasoning-traces}{allenai/big-reasoning-traces} & All & 2.44B & 1 \\
allenai/math-meta-reasoning-filtered & All & 1.24B & 1 \\
\href{https://huggingface.co/datasets/nvidia/OpenScience}{nvidia/OpenScience} & All & 9.7B & 2 \\
\bottomrule
\end{tabular}
\end{table}

\newpage
\subsection{Stage 3 (Stable+SqrtDecay) Data Mixture}

\textbf{Data composition:}

\begin{itemize}
    \item \textbf{Hindi text:} $\sim$53\% ($\sim$44B tokens)
    \item \textbf{Synthetic English text:} $\sim$35.6\% ($\sim$29B tokens)
    \item \textbf{Educational English text:} $\sim$0.4\% ($\sim$0.34B tokens)
    \item \textbf{Reasoning-focused English text}: $\sim$7\% ($\sim$6B tokens)
    \item \textbf{Math English text:} $\sim$4\% ($\sim$3B tokens)
    \item \textbf{Learning Rate:} starts at $7e-4$ and remains stable for the first 14,000 steps. It then decays in a negative square-root shape to a minimum learning rate ($0$) over the remaining 21,000 steps. The decay phase covers approximately 44 billion tokens, about $\sim$12\% of the total training tokens.
\end{itemize}

\begin{table}[h]
\centering
\caption{Data mixture for Stage-3}
\label{tab:stage3-hi}
\begin{tabular}{@{}l l c c@{}}
\toprule
\textbf{Dataset Name} & \textbf{Subset} & \textbf{Size (Tokens)} & \textbf{Repetition Factor} \\
\midrule
\href{https://huggingface.co/datasets/Polygl0t/gigalekh-v1}{\gigalekh} & Edu Score of 4 & 14.3B & 3 \\
 & Edu Score of 5 & 0.16B & 5 \\
\href{https://huggingface.co/datasets/HuggingFaceFW/fineweb-edu}{HuggingFaceFW/fineweb-edu} & Edu Score of 5 & 0.068B & 5 \\
\href{https://huggingface.co/datasets/HuggingFaceTB/smollm-corpus}{HuggingFaceTB/smollm-corpus} & All & 29.4B & 1 \\
\href{https://huggingface.co/datasets/HuggingFaceTB/finemath}{HuggingFaceTB/finemath} & Edu Score of 5 & 1.08B & 3 \\
\href{https://huggingface.co/datasets/allenai/big-reasoning-traces}{allenai/big-reasoning-traces} & All & 2.44B & 2 \\
allenai/math-meta-reasoning-filtered & All & 1.24B & 1 \\
\bottomrule
\end{tabular}
\end{table}
\newpage

\section{Evaluation Suite Analysis}
\label{appendix:evaluation-benchmarks}

\subsection{Motivation and Challenges}

Evaluations play a crucial role in monitoring model improvements during pretraining. However, for low-resource languages such as Hindi, many existing benchmarks are noisy and exhibit high variability.

To systematically assess which benchmarks provide meaningful signals of training progress, especially for smaller models like \lilmoo that are trained for shorter durations, we conducted an analysis inspired by \cite{penedo2025fineweb2}. This analysis focused on the first version of our monolingual Hindi model (\lilmoohi) and examined its respective language-specific benchmarks separately. The goal was to identify benchmarks that provide early indications of model improvement, which could then be used to compare performance between our models and baselines more fairly. Further details and results are provided in subsequent sections.

\subsection{Evaluation Suite Compilation}

To begin, we surveyed prior research that had already developed evaluation benchmarks for Hindi. Since reproducibility is a central concern in our work, we prioritized benchmarks that could be readily integrated into an automated evaluation pipeline (e.g., EleutherAI's Language Model Evaluation Harness \citep{eval-harness}). In total, we selected 10 different evaluations from different sources \cite{lai2023okapi, verma2024milu, kakwani2020indicnlpsuite, chang2025global}. Below, we list all of them.

\begin{itemize}
    \item \textbf{AI2 Reasoning Challenge (ARC):} The ARC dataset is a collection of multiple-choice questions designed to test a model's ability to reason and answer questions that require more than just surface-level understanding.
    
    \item \textbf{Massive Multitask Language Understanding (MMLU):} MMLU is a benchmark that evaluates a model's performance across a wide range of Academic subjects, testing its ability to understand and reason about complex topics.
    
    \item \textbf{HellaSwag:} HellaSwag is a benchmark for commonsense reasoning and natural language understanding. The task is to select the most plausible continuation of a given story or scenario from multiple options.  
    
    \item \textbf{TruthfulQA:} A benchmark to measure whether a language model is truthful in generating answers to questions. The benchmark comprises multiple questions across categories such as health, law, finance, and politics.
    
    \item \textbf{Global PIQA}: Global PIQA is a benchmark for physical commonsense reasoning in multiple languages, including Hindi. The task is to select the most plausible solution to a given physical problem from two options.
    
    \item \textbf{Multi-task Indic Language Understanding Benchmark (MILU):} It is a large-scale evaluation suite covering 11 Indic languages, 8 domains, and 41 subjects. It is designed to assess both general knowledge and culturally grounded understanding relevant to the Indian context.   
    
    \item \textbf{Cloze-style Multiple Choice QA (CSQA):} is a question answering task in which an entity in a sentence is masked, and the model must select the correct entity from four candidates, emphasizing commonsense and relational reasoning. 
    
    \item \textbf{Choice of Plausible Alternatives (COPA):} tests open-domain causal reasoning. Given a premise, models must identify the more plausible cause or effect from two alternatives, requiring an understanding of causal relationships.
    
    \item \textbf{IIT-Patna Movie Review (IITP-MR):} is a sentiment analysis dataset annotated with three classes---positive, negative, and neutral---used to evaluate sentiment understanding in Hindi text.
    
    \item \textbf{IndicXNLI:} is a Natural Language Inference benchmark for 11 Indic languages, created via translation of the English XNLI dataset. It evaluates a model’s ability to determine whether a pair of sentences is entailed, contradictory, or neutral.
\end{itemize}

These benchmarks can be categorized into two common categories: Multiple-Choice Format (MCF) and Cloze Format (CF). In MCF tasks (e.g., MMLU, MILU, ARC), the model is presented with all candidate answer options explicitly in the prompt, typically labeled as A/B/C/D or option1/option2, and is required to select the correct choice (i.e., most likely). In contrast, CF tasks (e.g., HellaSwag, Global PIQA, CSQA) do not present answer choices directly. Several studies \citep{gu2025olmes,li2024datacomp,penedo2025fineweb2} have shown that models often struggle with MCF tasks in the early stages of training, with performance improving only after extensive training or exposure to substantially larger datasets. As a result, MCF benchmarks may be less sensitive to improvements in small-scale models, such as those developed in this work. Conversely, CF evaluations tend to provide more reliable early signals and are better suited for assessing models trained with limited data.

Table~\ref{tab:eval-harness} summarizes the benchmarks, including their n-shot settings, baselines, and evaluation metrics. All benchmarks(except TruthfulQA) were assessed using a 5-shot configuration.

\begin{table}[h]
\centering
\caption{Summary of Evaluation Benchmarks}
\label{tab:eval-harness}
\begin{tabular}{@{}l r r l@{}}
\toprule
\textbf{Benchmark} & \textbf{n-shot} & \textbf{Baseline} & \textbf{Metric} \\
\midrule
ARC & 5-shot & 25 & \texttt{acc} \\
MMLU & 5-shot & 25 & \texttt{acc} \\
HellaSwag & 5-shot & 25 & \texttt{acc} \\
TruthfulQA & 0-shot & 22.5 & \texttt{mc1} \\
MILU & 5-shot & 25 & \texttt{acc} \\
CSQA & 5-shot & 25 & \texttt{acc} \\
Global PIQA & 5-shot & 50 & \texttt{acc} \\
COPA & 5-shot & 50 & \texttt{acc} \\
IITP-MR & 5-shot & 33.3 & \texttt{acc} \\
IndicXNLI & 5-shot & 33.3 & \texttt{acc} \\
\bottomrule
\end{tabular}
\end{table}

\subsection{Per-benchmark Performance}

\begin{figure}[H]
\centering

\begin{subfigure}[b]{0.48\textwidth}
    \centering
    \includegraphics[width=0.9\textwidth]{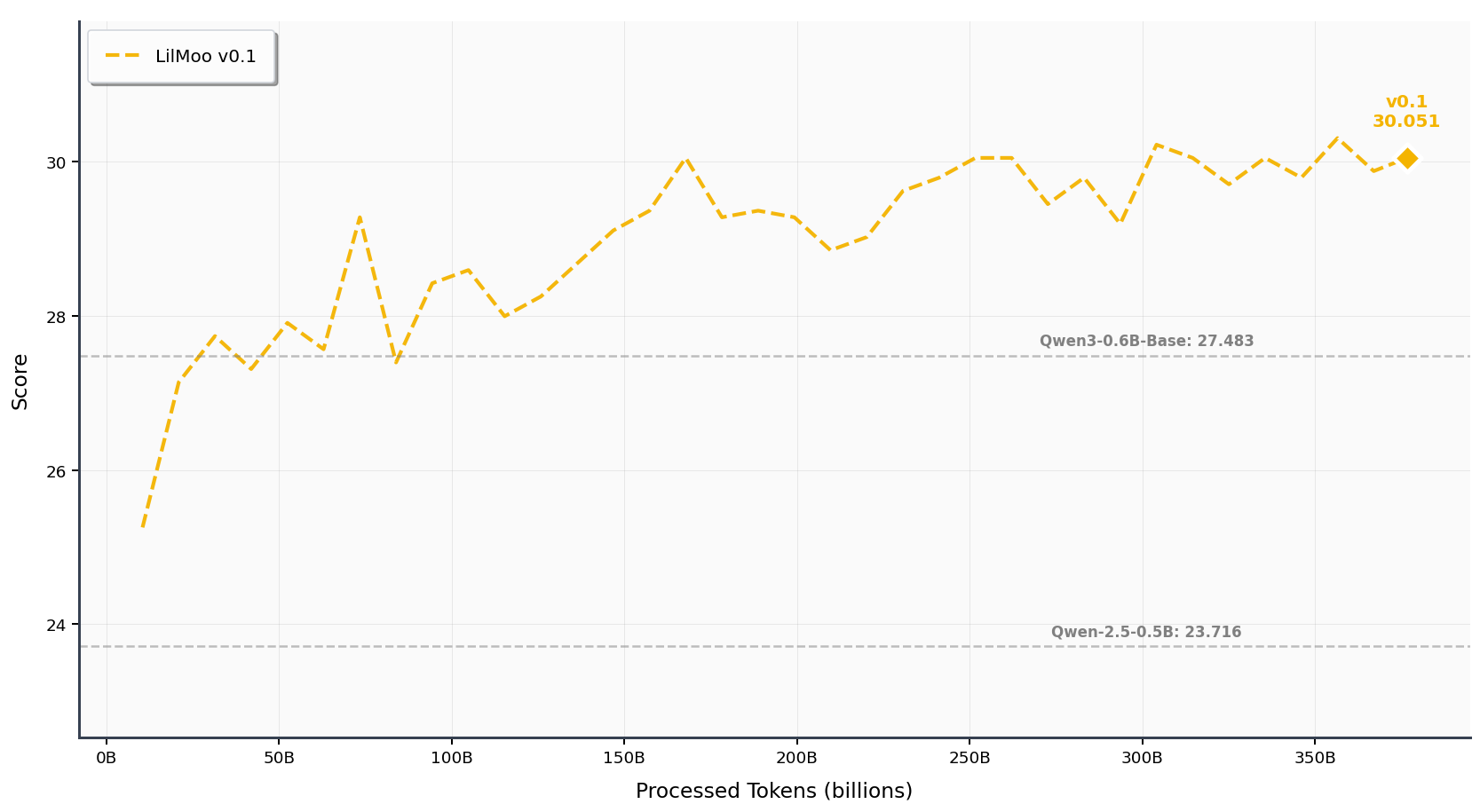}
    \caption{ARC}
\end{subfigure}
\hfill
\begin{subfigure}[b]{0.48\textwidth}
    \centering
    \includegraphics[width=0.9\textwidth]{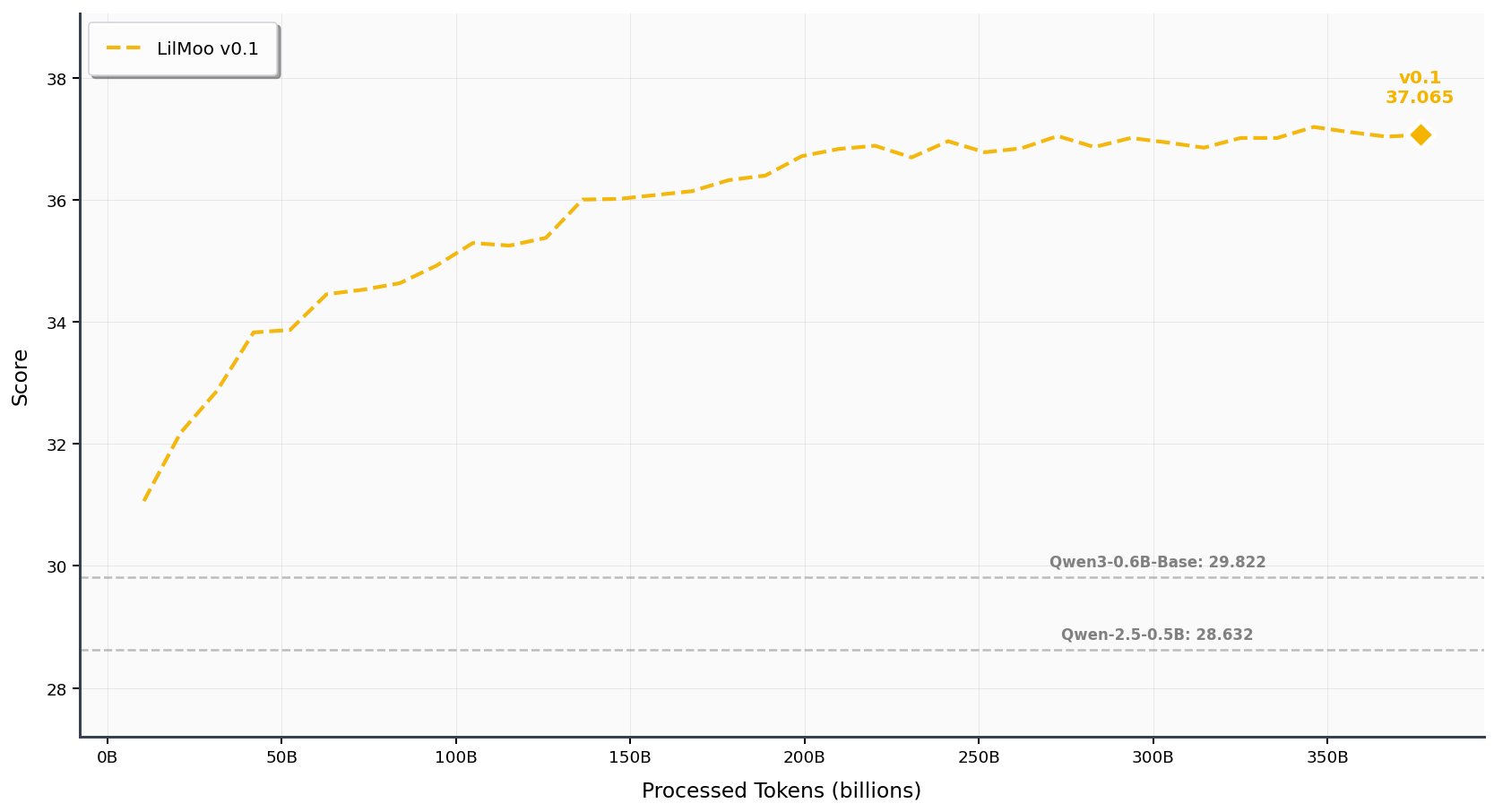}
    \caption{Hellaswag}
\end{subfigure}

\vspace{0.3cm}

\begin{subfigure}[b]{0.48\textwidth}
    \centering
    \includegraphics[width=0.9\textwidth]{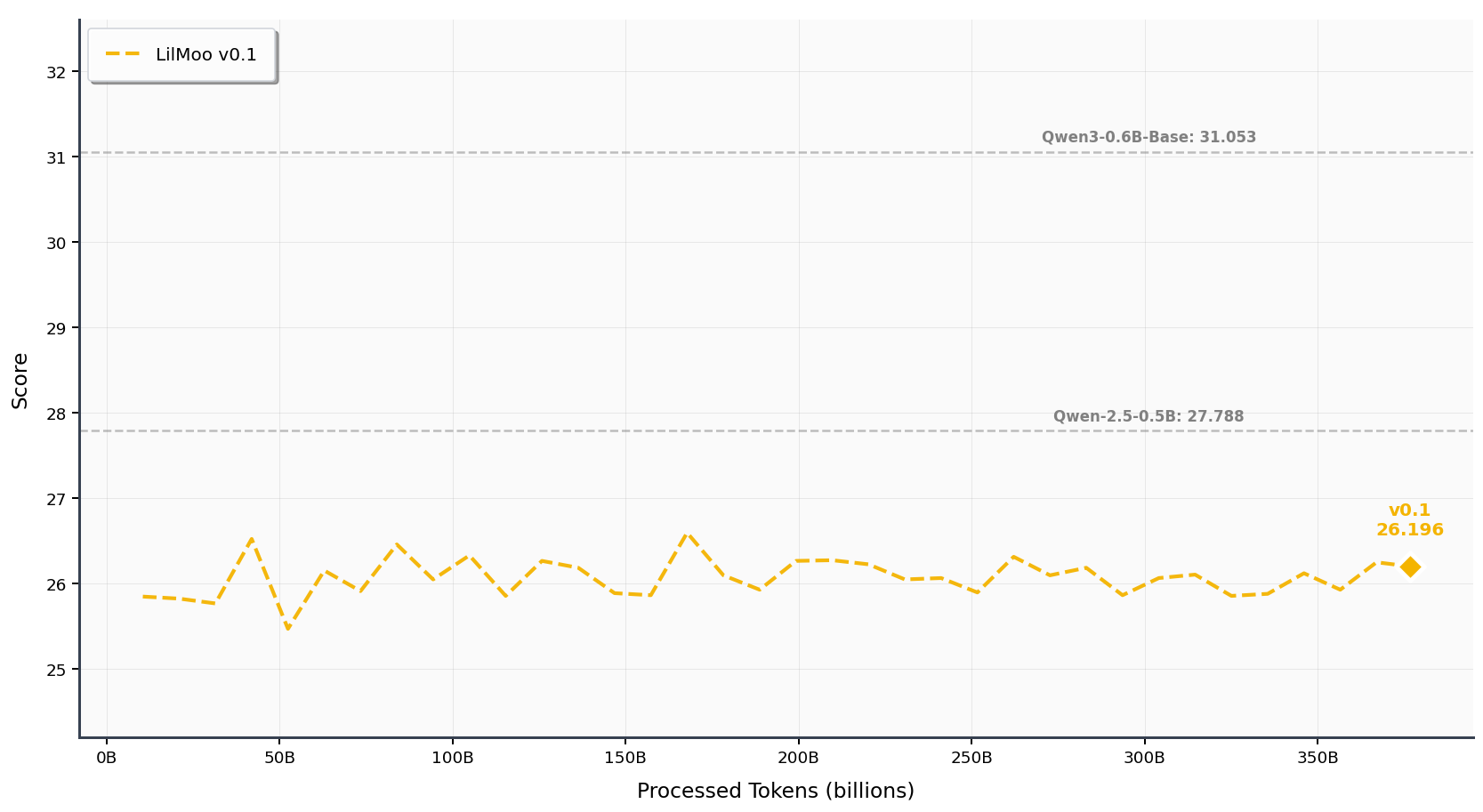}
    \caption{MMLU}
\end{subfigure}
\hfill
\begin{subfigure}[b]{0.48\textwidth}
    \centering
    \includegraphics[width=0.9\textwidth]{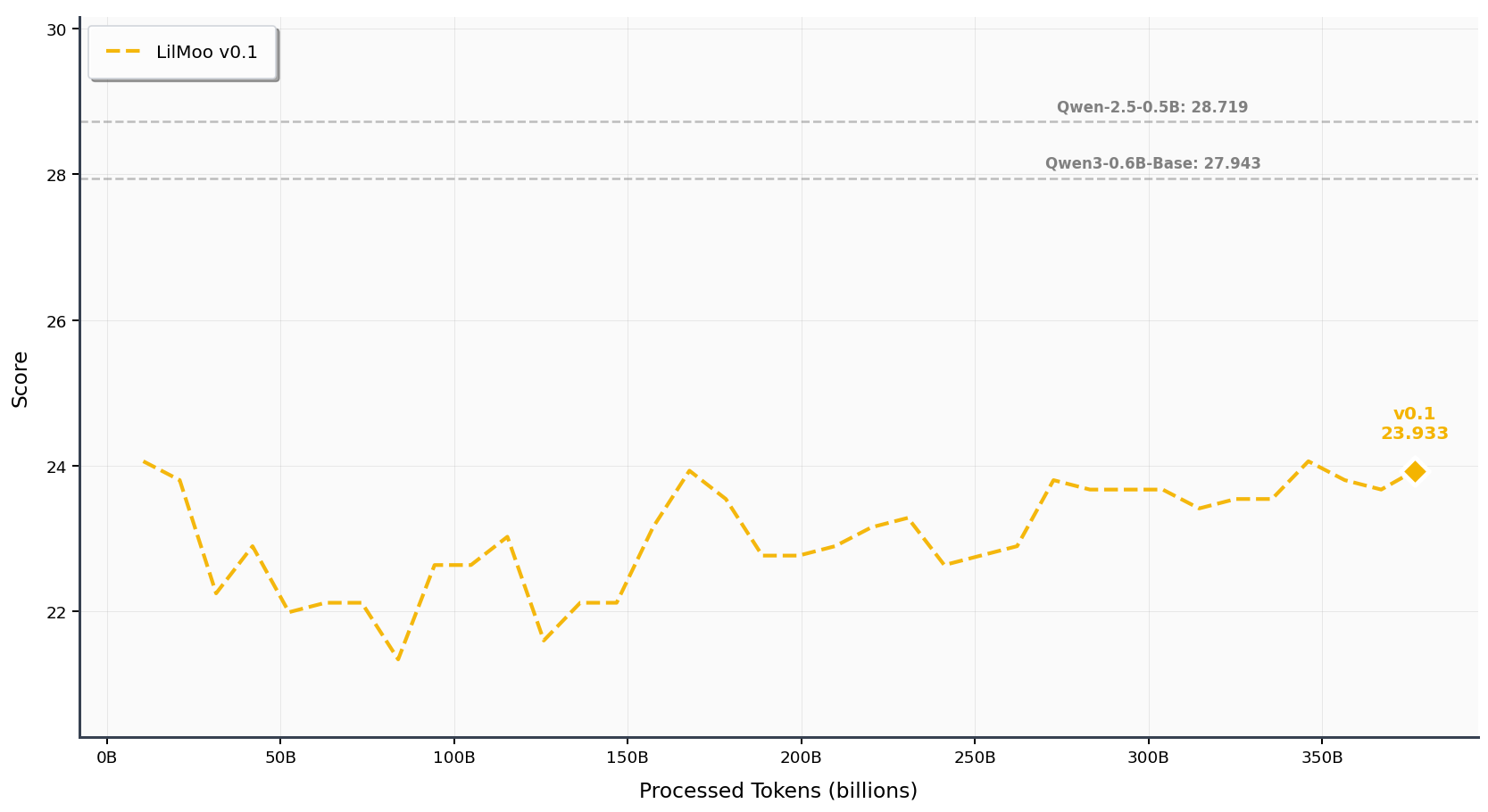}
    \caption{TruthfulQA}
\end{subfigure}

\vspace{0.3cm}

\begin{subfigure}[b]{0.48\textwidth}
    \centering
    \includegraphics[width=0.9\textwidth]{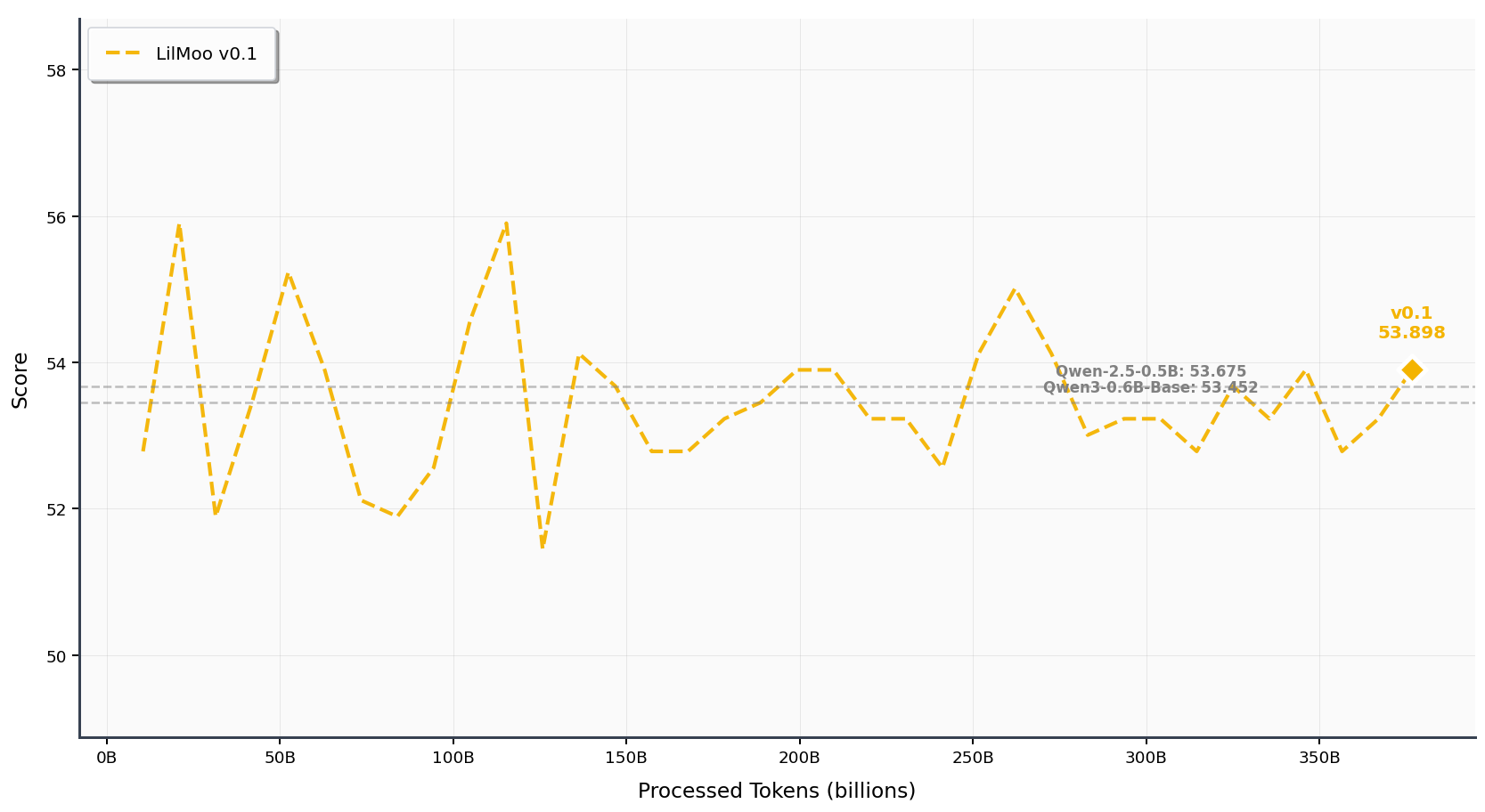}
    \caption{COPA}
\end{subfigure}
\hfill
\begin{subfigure}[b]{0.48\textwidth}
    \centering
    \includegraphics[width=0.9\textwidth]{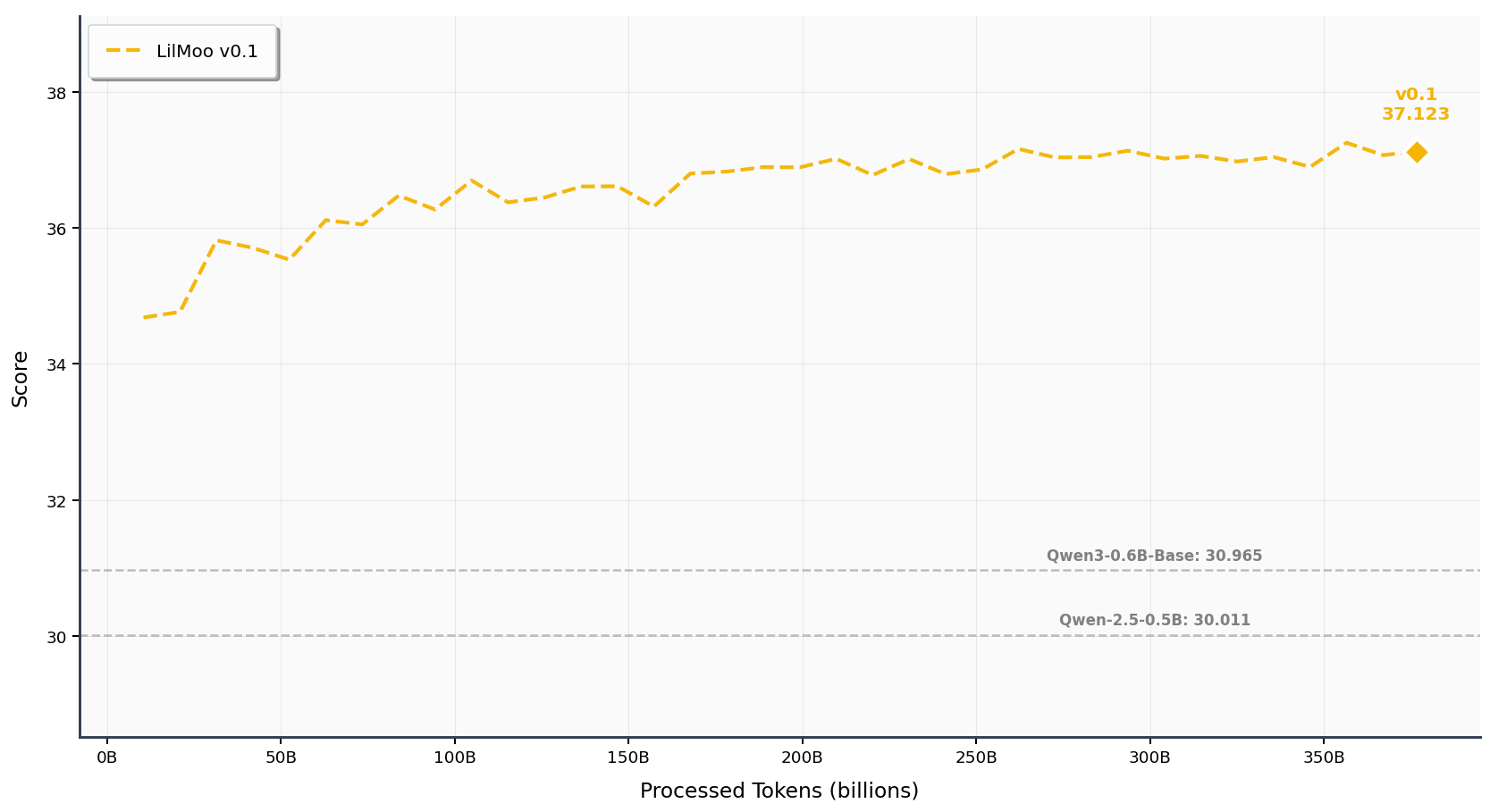}
    \caption{CSQA}
\end{subfigure}

\vspace{0.3cm}

\begin{subfigure}[b]{0.48\textwidth}
    \centering
    \includegraphics[width=0.9\textwidth]{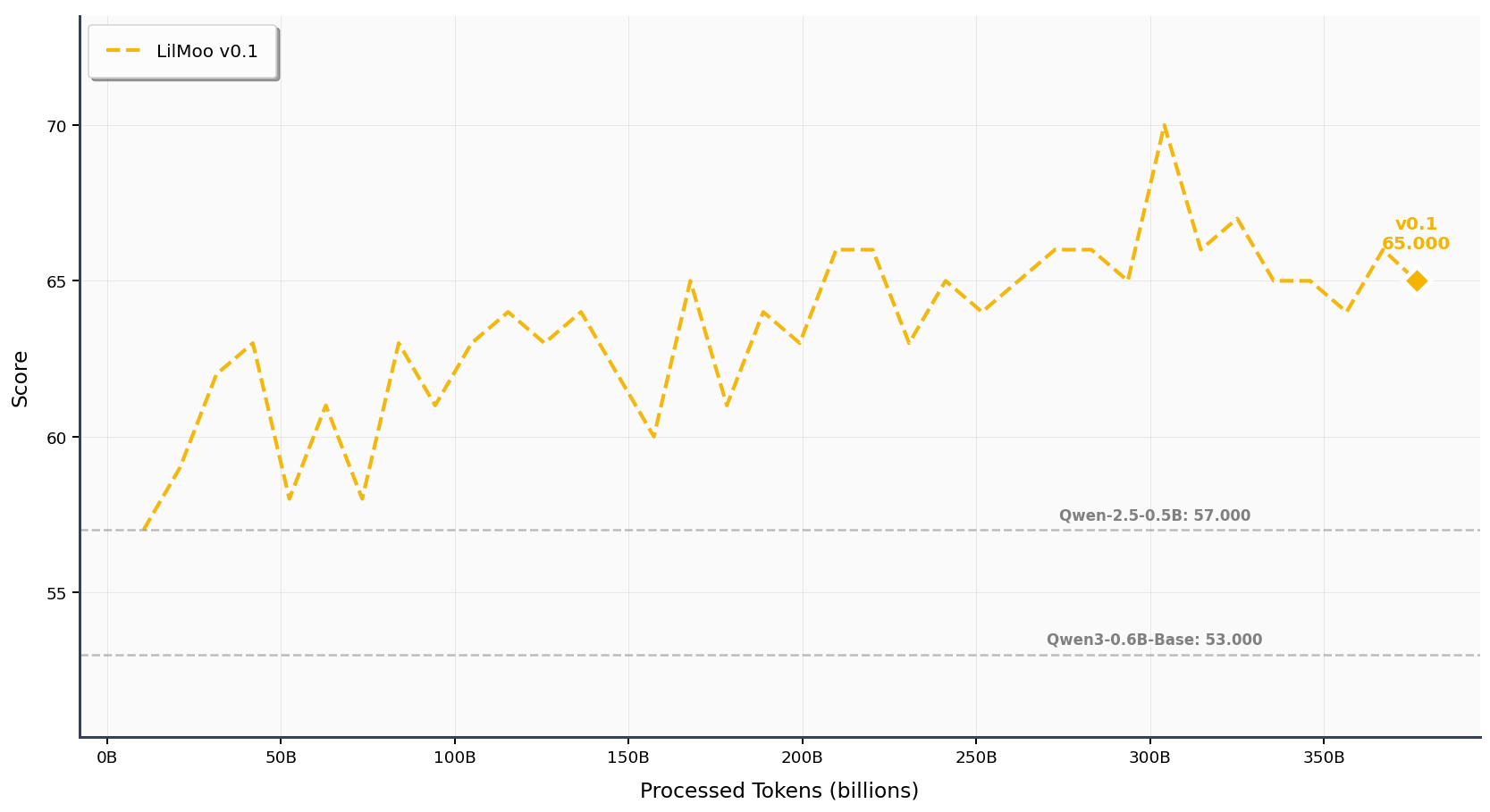}
    \caption{Global PIQA}
\end{subfigure}
\hfill
\begin{subfigure}[b]{0.48\textwidth}
    \centering
    \includegraphics[width=0.9\textwidth]{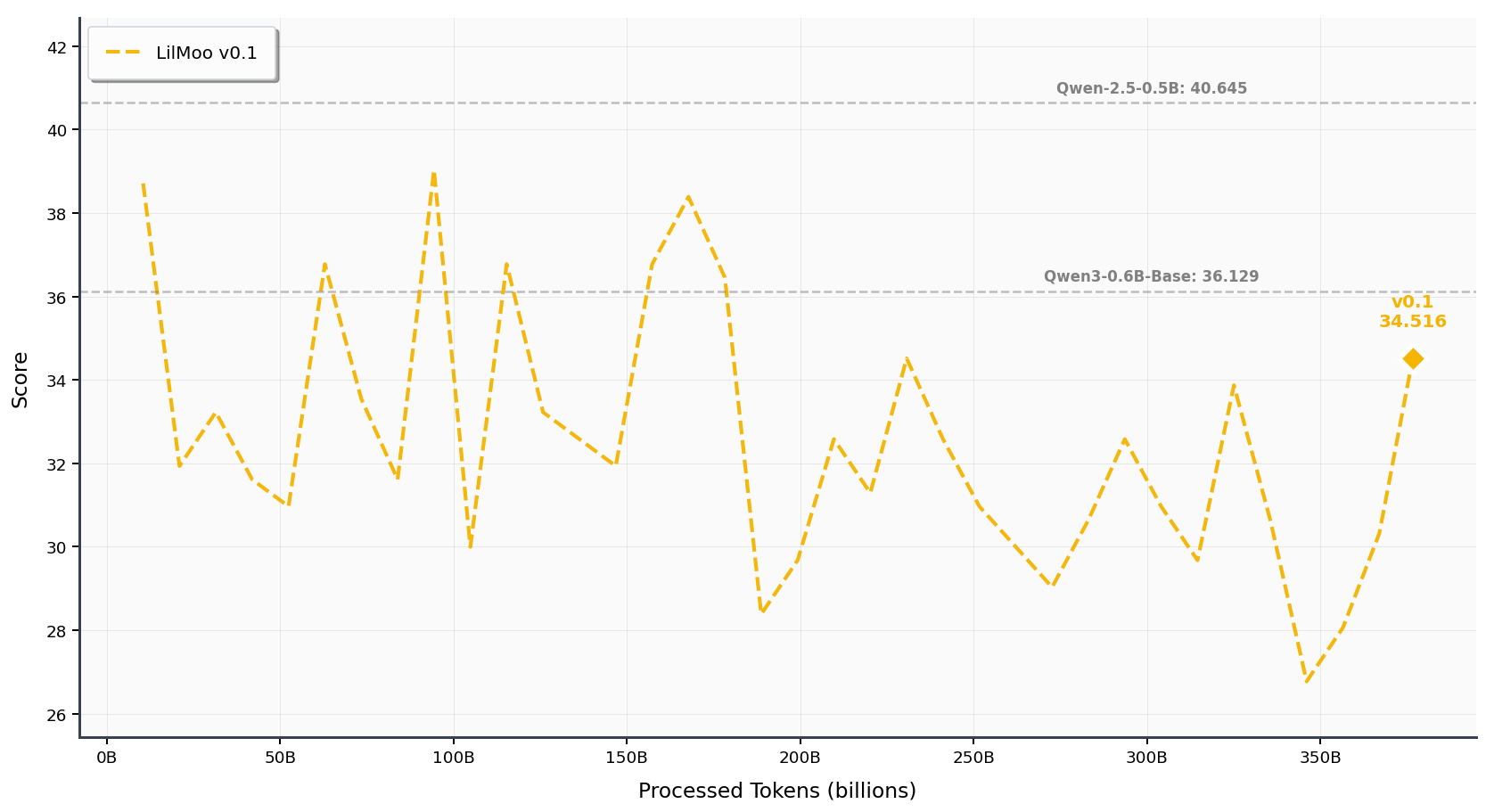}
    \caption{IITP\_MR}
\end{subfigure}

\vspace{0.3cm}

\begin{subfigure}[b]{0.48\textwidth}
    \centering
    \includegraphics[width=0.9\textwidth]{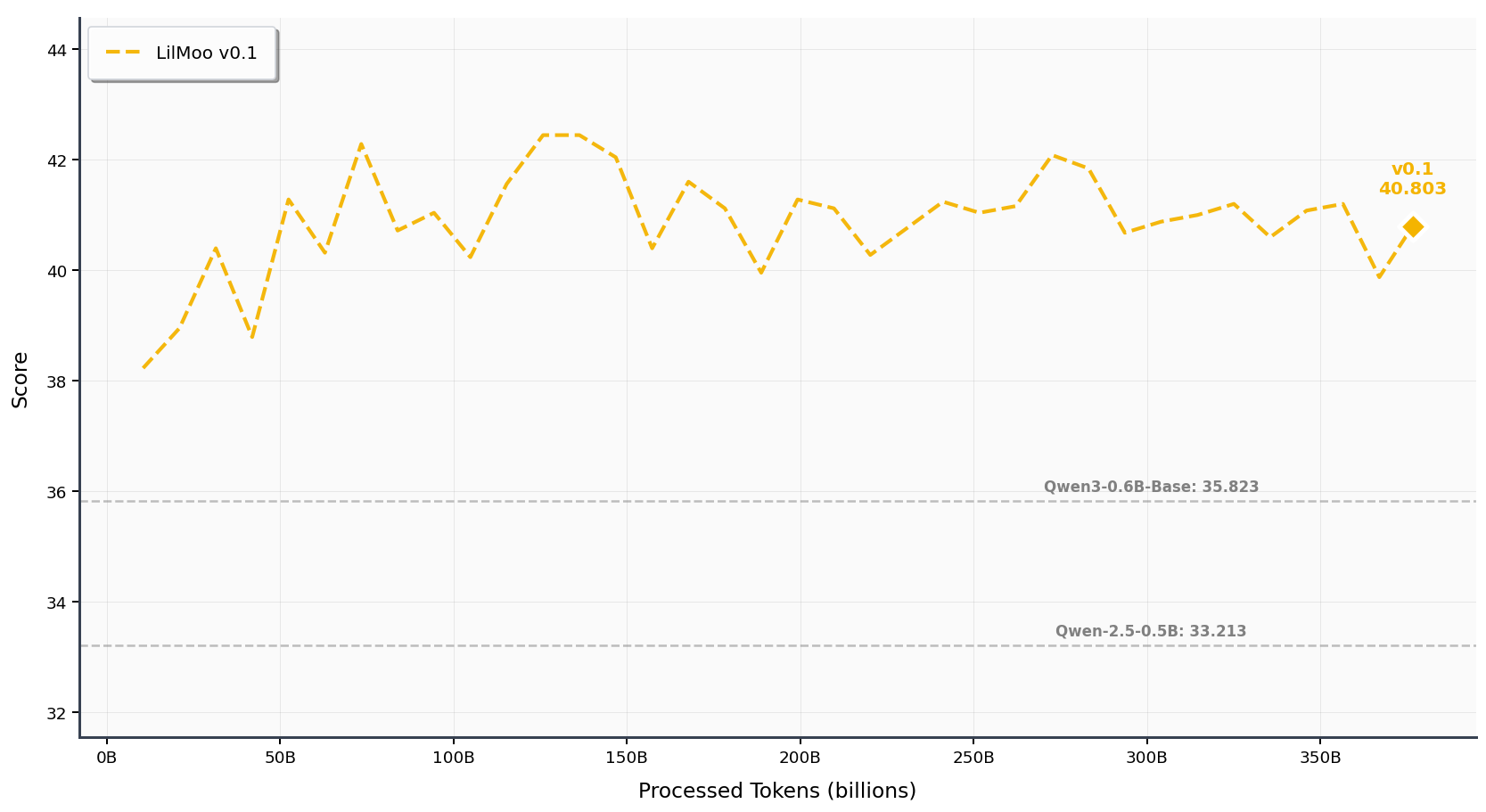}
    \caption{IndicXNLI}
\end{subfigure}
\hfill
\begin{subfigure}[b]{0.48\textwidth}
    \centering
    \includegraphics[width=0.9\textwidth]{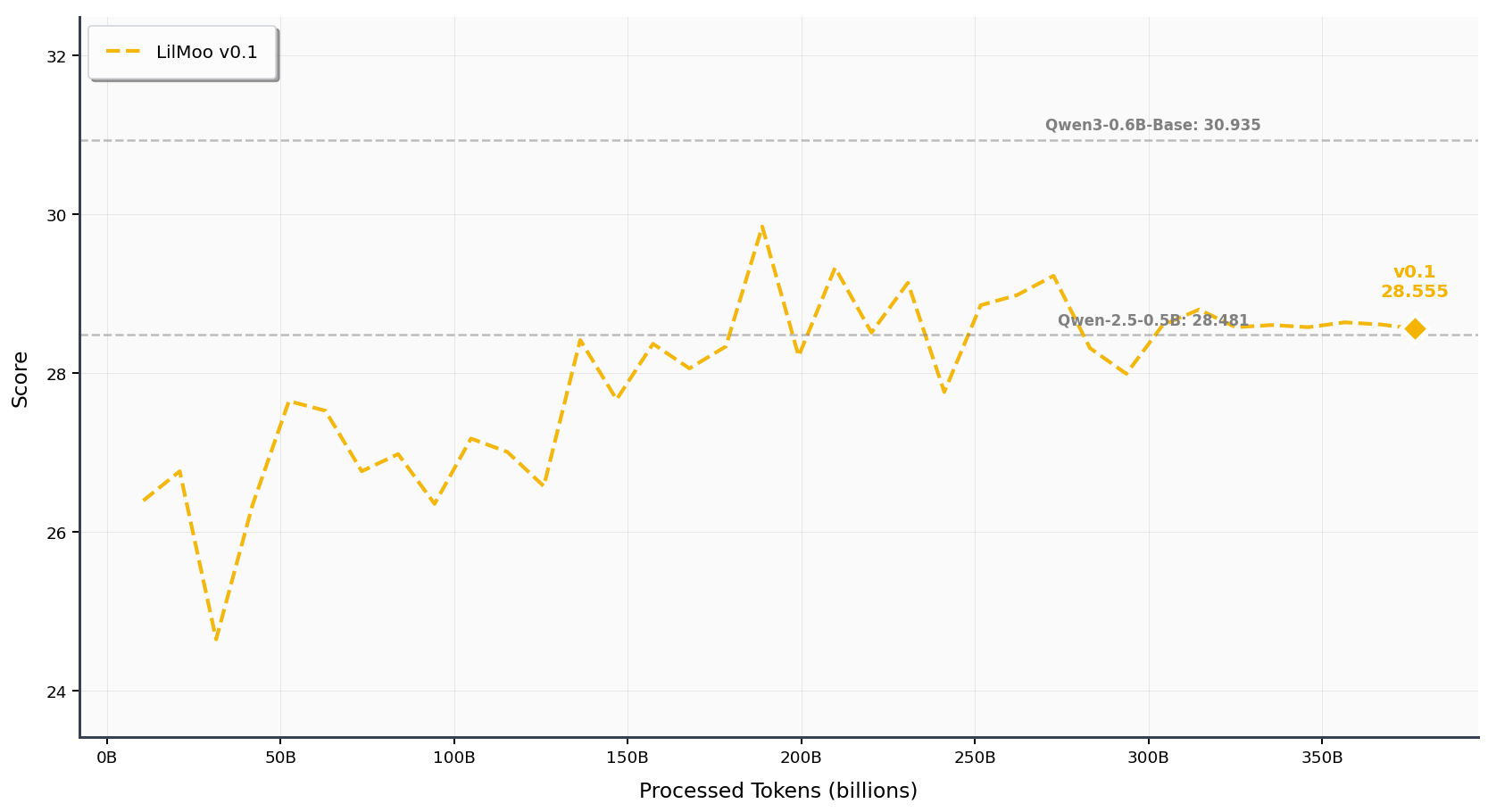}
    \caption{MILU}
\end{subfigure}

\caption{Per-benchmark Performance across training checkpoints.}
\label{fig:benchmark-al-lilmoov1}
\end{figure}

Figure~\ref{fig:benchmark-al-lilmoov1} plots the performance of \lilmoohi during pretraining on the previously described evaluation suite. The figures show performance changes across the benchmarks as the training data increases. Every data point corresponds to a checkpoint saved mid-training ($\sim$10 billion tokens). We also add some baselines (Qwen models) for reference.

\newpage

ARC, HellaSwag, and CSQA show clear and consistent improvement across training, with only minor fluctuations. The consistent upward trend indicates that these benchmarks are sensitive to gains in reasoning and comprehension as pretraining progresses. Although Global PIQA and MILU exhibit noticeable fluctuations across checkpoints, both benchmarks still display an overall upward trend; in particular, later checkpoints consistently outperform earlier ones. IndicXNLI shows an upward trend early in training but plateaus later. The remaining benchmarks, TruthfulQA, MMLU, IITP-MR, and COPA, exhibit more erratic behavior, with performance fluctuating substantially. This inconsistency raises questions about their reliability as indicators of model improvement during pretraining. 

\subsection{Aggregate Performance (Normalized Preferred Metric)}
\label{appendix:npm}

To offer a single performance summary across all benchmarks, we adopted the Normalized Preferred Metric (NPM) proposed by \cite{pires2023sabia}. The NPM is an aggregation metric that provides an overall performance summary by normalizing each task score with respect to its expected random and maximum scores. The NPM is calculated as follows:

$$
\text{NPM} = \frac{1}{N} \sum_{i=1}^{N} 100 \times \frac{\text{Preferred Metric}_i - \text{Random Score}_i}{\text{Max Score}_i - \text{Random Score}_i}
$$

\begin{figure}[h]
\centering
\includegraphics[width=\linewidth]{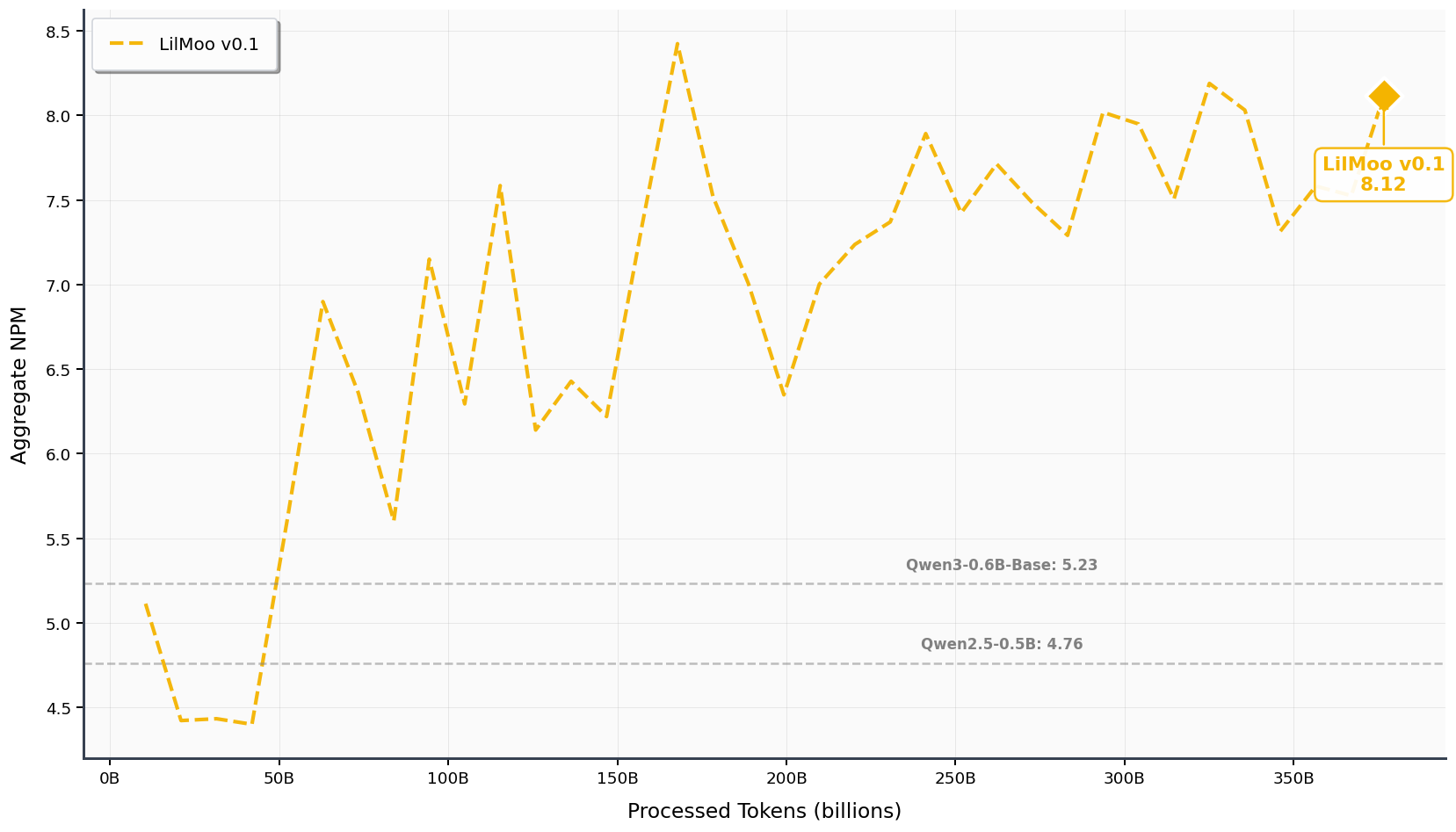}
\caption{NPM aggregated score of all benchmarks.}
\label{fig:npm-lilmoov1}
\end{figure}

Figure~\ref{fig:npm-lilmoov1} plots the performance of \lilmoohi on NPM of all benchmarks. The aggregate NPM steadily increases with training, indicating that \lilmoohi makes consistent overall progress across the evaluated benchmarks as pretraining advances. Despite some fluctuations, the model surpasses the reference baselines. 

\subsection{Baseline Surpassing Analysis}

Table~\ref{tab:baseline-surpass-hi} shows the number of training tokens required for \lilmoohi to surpass each benchmark’s baseline by 5\% points. The results show a separation between benchmarks that provide an early and strong learning signal and those that remain challenging throughout pretraining. Several benchmarks are surpassed very early in training, while others are not crossed at all within the observed training window.

\begin{table}[!htbp]
\centering
\small
\caption{Training tokens required for v0.1 to surpass random baselines by 5\%.}
\label{tab:baseline-surpass-hi}
\begin{tabular}{lrrrrrr}
\toprule
\textbf{Benchmark} & \textbf{Baseline} & \textbf{Threshold} & \textbf{First Step} & \textbf{Tokens (B)} & \textbf{Score} & \textbf{\%} \\
\midrule
ARC & 0.250 & 0.300 & 80000 & 167.770 & 0.301 & +0.051 \\
HellaSwag & 0.250 & 0.300 & 5000 & 10.490 & 0.311 & +0.061 \\
MMLU & 0.250 & 0.300 & Not Reached & - & - & - \\
TruthfulQA MC1 & 0.225 & 0.275 & Not Reached & - & - & - \\
COPA & 0.500 & 0.550 & 10000 & 20.970 & 0.559 & +0.059 \\
IITP MR & 0.330 & 0.380 & 5000 & 10.490 & 0.387 & +0.057 \\
IndicXNLI & 0.330 & 0.380 & 5000 & 10.490 & 0.382 & +0.052 \\
MILU & 0.250 & 0.300 & Not Reached & - & - & - \\
CSQA & 0.250 & 0.300 & 5000 & 10.490 & 0.347 & +0.097 \\
Global PIQA & 0.500 & 0.550 & 5000 & 10.490 & 0.570 & +0.070 \\
\bottomrule
\end{tabular}
\vspace{0.3cm}
\begin{flushleft}
\small
\end{flushleft}
\end{table}

Among the earliest benchmarks to be surpassed, CSQA, Global PIQA, and IITP-MR crossed their respective thresholds at the first evaluation checkpoint (5k steps, $\sim$10B tokens). CSQA shows a particularly strong margin over the baseline. MMLU, TruthfulQA, and MILU are not surpassed within the observed training window. These benchmarks likely demand substantially more training data, higher model capacity, or more targeted pretraining data, and therefore appear better suited for evaluating later stages of training or larger model scales.

\subsection{Noise Analysis}

Evaluation results from a benchmark suite can be interpreted as a signal of how well a model is learning. When this signal is strong and consistent, it indicates that the benchmark reliably captures the model's capabilities. Conversely, weak or noisy signals make it difficult to derive meaningful insights regarding model performance. To assess the stability and reliability of each benchmark's signal, we computed several statistics quantifying evaluation noise across checkpoints. Following the methodology of \cite{penedo2025fineweb2}, we employ the following metrics:

\begin{itemize}
\item \textbf{Mean Absolute Change (MAC):} The average absolute difference between successive evaluation points, providing a measure of step-to-step variability.
\item \textbf{Signal-to-Noise Ratio (SNR):} The ratio of the mean benchmark score to its standard deviation. Higher values indicate that the underlying signal dominates random fluctuations.
\item \textbf{Spearman Correlation:} The correlation between benchmark scores and training steps, where higher positive correlations suggest that the benchmark effectively tracks learning progression.
\end{itemize}

\begin{figure}[H] 
\centering

\begin{subfigure}[b]{0.48\textwidth}
    \centering
    \includegraphics[width=\textwidth]{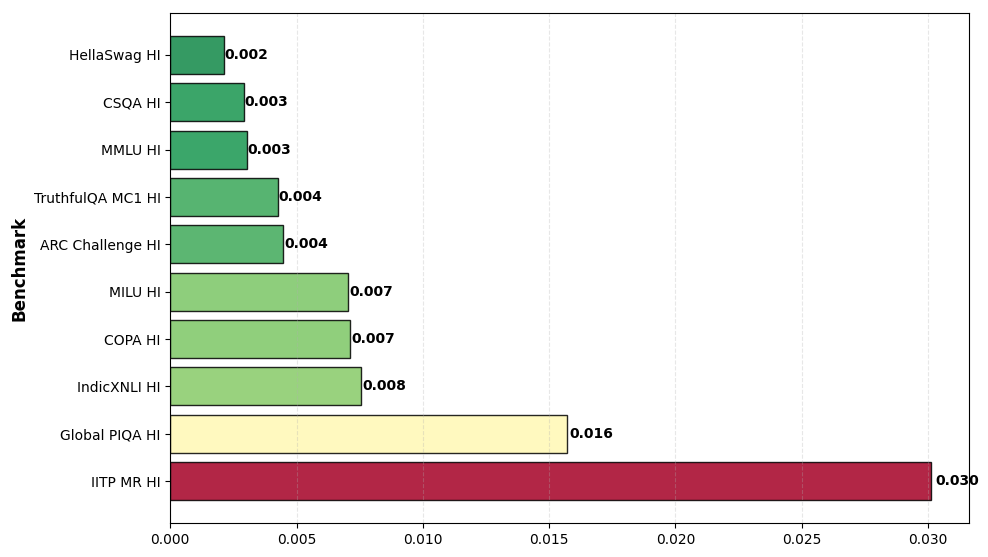}
    \caption{Mean Absolute Change}
    \label{fig:noise-mean-abs-change}
\end{subfigure}
\hfill
\begin{subfigure}[b]{0.48\textwidth}
    \centering
    \includegraphics[width=\textwidth]{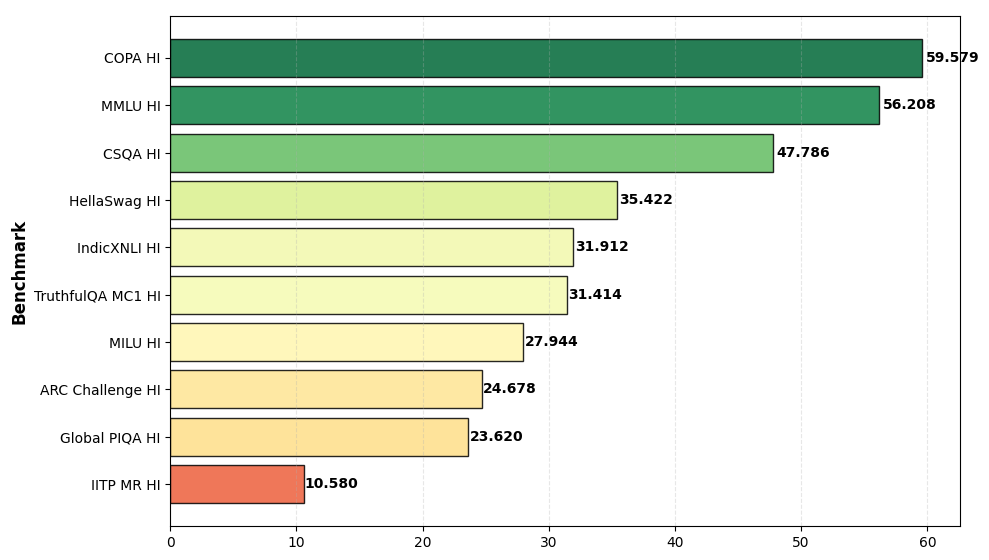}
    \caption{Signal-to-Noise Ratio}
    \label{fig:noise-snr}
\end{subfigure}

\vspace{0.5cm}

\begin{subfigure}[b]{0.48\textwidth}
    \centering
    \includegraphics[width=\textwidth]{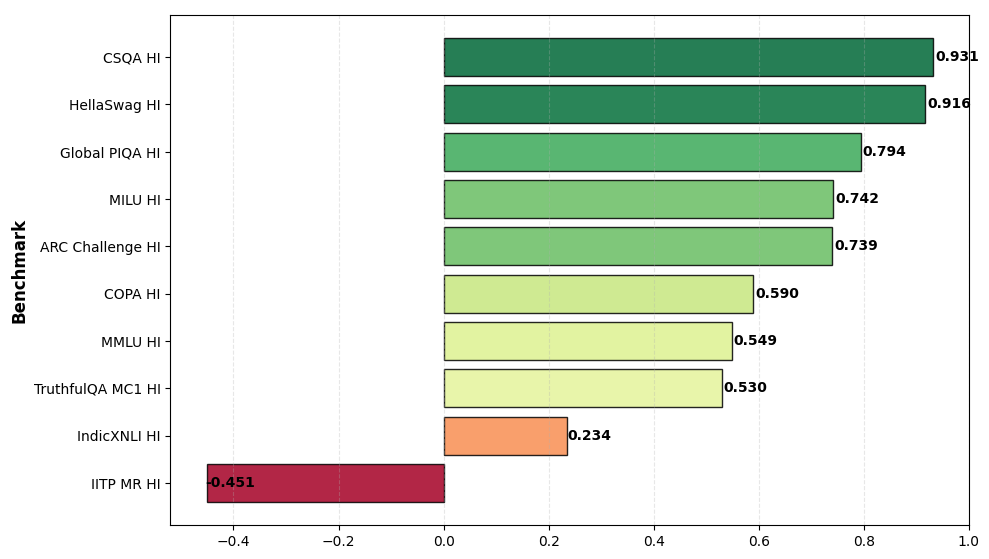}
    \caption{Spearman Correlation}
    \label{fig:noise-spearman}
\end{subfigure}

\caption{Noise analysis for the 10 benchmarks in our Hindi evaluation suite.}
\label{fig:noise-lilmoov1}
\end{figure}

Figure~\ref{fig:noise-lilmoov1} presents the three signal-quality metrics across all benchmarks. 

\newpage

From these results, we can see that:

\begin{itemize}

\item Most benchmarks exhibit strong positive Spearman Correlation $(>0.5)$ with training steps. In particular, CSQA and HellaSwag exhibit strong monotonic alignment with training progress, while Global PIQA, MILU, and ARC also demonstrate this behavior. The main exception is IITP-MR, which shows a negative correlation.

\item Score volatility is generally low to moderate across most benchmarks, with Mean Absolute Changes remaining below 0.01 for most of the benchmarks. In contrast, IITP-MR and Global PIQA show noticeably higher volatility.

\end{itemize}

After this analysis, we selected benchmarks that reliably track model improvement over time and dropped those that exhibited high noise or poor alignment with training progress.

The benchmarks retained---HellaSwag, CSQA, ARC, MMLU, MILU, and Global PIQA---were selected because they reliably track model improvement during pretraining. They exhibit high Spearman correlations, low-to-moderate volatility, and strong Signal-to-Noise Ratios, producing smooth, monotonic checkpoint curves that clearly reflect learning progress. Even Global PIQA, with higher score variability, maintains a strong positive correlation with training, making it useful for monitoring. Meanwhile, benchmarks such as TruthfulQA, COPA, IndicXNLI, and IITP-MR were excluded because they exhibit low correlation with model improvement, high volatility, or poor SNR.

\end{document}